\documentclass[final]{hubsarxiv}

\usepackage[per-mode=symbol]{siunitx}
\usepackage{longtable}
\usepackage{pgfplots}
\usepackage{pgfplotstable}
\pgfplotsset{compat=1.18}
\usepgfplotslibrary{dateplot}
\usetikzlibrary{positioning,arrows.meta,fit,backgrounds}
\usepackage{listings}
\AtBeginEnvironment{lstlisting}{\linespread{1}\selectfont}
\AtBeginEnvironment{table}{\setlength{\belowcaptionskip}{6pt}}

\newcommand{\CI}{C_{\mathrm{ans}}(I)}
\newcommand{\Cact}{C_{\mathrm{act}}(I)}
\newcommand{\Acc}{A}
\newcommand{\Areal}{A_{\mathrm{r}}}
\newcommand{\dlim}{D}
\newcommand{\aifail}{F}
\newcommand{\unp}{U}
\newcommand{\cS}{c_{\Sigma}}
\definecolor{creal}{HTML}{2C7BB6}
\definecolor{caifail}{HTML}{D7191C}
\definecolor{cdata}{HTML}{BDBDBD}
\definecolor{cb}{HTML}{2C7BB6}\definecolor{cr}{HTML}{D7191C}

\title{Which Buildings Are Artificial Intelligence-Ready? A Measurement-Based Assessment Framework for AI Question Answering and Actuation}
\shorttitle{Which buildings are AI-ready?}
\shortauthors{Wooyoung Jung}
\author{Wooyoung Jung\affmark{a,}\corrmark\orcid{0000-0003-2746-4983}}
\affil{\affmark{a}Department of Civil and Architectural Engineering and Mechanics, University of Arizona, Tucson, AZ, USA}
\corresponding{wooyoung@arizona.edu}
\paperno{2026-008}
\paperdate{October 2026}
\versionnote{v1 (2026-10-06): first release. v2 (2026-11-05): re-set in the HUBS Lab paper layout; content unchanged.}
\license{CC BY-NC-ND 4.0}
\abstract{Agentic artificial intelligence (AI) systems are becoming the interface to buildings, answering
questions and controlling operations, but a building's readiness for them has not been
systematically assessed. This study proposes a framework to quantify it. First, a building's
knowledge graph sets two ceilings. The answerable-readiness ceiling is the share of operational
questions its data could answer, and the actuation-readiness ceiling is the share of control actions
it exposes. Second, a reference AI agent's accuracy on a fixed set of these questions shows how much
of the ceilings is realized. On a simulated office, the agent realizes 0.62 of a 0.64 answerable
ceiling, so missing data, not the AI, limit readiness, except in naming a fault's cause. Across 37
public real-building graphs, the median answerable ceiling is 0.16, and in 15 of 45, unlinked
sensors lower it. The framework turns ``is this building AI-ready?'' into an auditable, ranked
retrofit question.}

\begin{document}
\maketitle

\section{Introduction}
\label{sec:intro}

The prospect of buildings managed with artificial intelligence
(AI)~\cite{bommasani2021foundation,react2023}, able to answer a user's questions in natural
language, diagnose faults, and eventually close control loops, has moved from speculation to active
engineering~\cite{liu2025llmbuilding,arslan2026llmsurvey}. Two of these tasks, fault diagnosis and
control, already reach real buildings without such AI. Fault detection is commercial practice, with
analytics platforms running rule-based checks over building automation data at
scale~\cite{clockworks2026,skyspark2026}. Learned and model-based control has been tested in field
studies. A reinforcement-learning policy controlled a conference room's supply airflow for three
weeks~\cite{chen2019gnurl}, and model predictive control ran an office's heating, ventilation and
air-conditioning (HVAC) for two months~\cite{blum2022mpcfield}. A growing number of studies now put
large language models (LLMs) in these roles. LLMs answer operational questions posed in natural
language over a building's sensor streams~\cite{lee2025hvacqa}, diagnose HVAC faults and explain
them to operators~\cite{zhang2025hvacfdd}, issue control actions in simulated
buildings~\cite{song2023industrialcontrol}, translate natural-language instructions into
building-management actions~\cite{jobe2026intent}, assist building energy
modelling~\cite{zhang2024llmbem}, and drive agentic workflows for building energy
analysis~\cite{zhang2025llmagentbuilding}.

To date, most of this research effort is directed at the AI: which model to use, which prompting
technique~\cite{jung2026cot}, how well the model retrieves and reasons, and what a new AI-integrated
system looks like~\cite{arslan2026llmsurvey,liu2025llmbuilding,jung2026hema}. This focus is natural,
because the field is young, AI capabilities change quickly, and the building community needs to
understand what the AI can do. But building-AI integration has two sides, and the building's data
infrastructure is the other. That infrastructure is the building's \emph{knowledge graph} (KG),
which describes its zones, equipment and points and the relations among them, such as which air
handler serves which zone, in a machine-readable schema such as Brick, together with the
operational, reference and control data the KG exposes. The AI operates on that data, and the data
bound what the AI can do however good the AI is. An AI cannot report a zone's energy use if the zone
is not metered, and it cannot attribute a comfort complaint to a fouled coil if the coil's behaviour
is never sensed. This raises a prior question: given a particular building, can its data answer the
questions that would be put to such a system? This study calls this property a building's
\emph{AI-readiness}: the extent to which its data infrastructure can answer the operational
questions stakeholders ask and expose the controls to act on them. AI-readiness is upstream of
deploying any AI system, so it can be assessed before any model is chosen, yet the assessment tools
available today measure something adjacent to it. For example, the European Union (EU) Smart
Readiness Indicator (SRI)~\cite{verbeke2020sri,ec2020sri} and SmartScore, the certification of the
company WiredScore~\cite{wiredscore2026}, are expert checklists that credit a building for the
services it has installed. Semantic-data standards such as Brick and Standard~223P of the American
Society of Heating, Refrigerating and Air-Conditioning Engineers
(ASHRAE)~\cite{balaji2018brick,ashrae223p} describe a building's points and equipment but mandate no
sensing. Building question-answering (QA) benchmarks test an AI over the KG, but they hold the
buildings fixed and grade the AI. Application shapes check whether a KG declares the points an
application needs, not whether the series behind them are usable. None measures what a building's
data can answer, and none separates a building's data limits from the AI's shortfall
(Section~\ref{sec:related}).

The goal of this study is to define and measure a building's AI-readiness as a property of its
instrumentation, so that ``is this building AI-ready?'' turns from a judgement into an auditable
assessment. Four constructs, formalized in Section~\ref{sec:framework}, carry the measurement:
\begin{itemize}
\item \emph{Question canon}: a declared question set formed by building domains, question
  intents, and answer-completeness levels;
\item \emph{Answerable-readiness ceiling}: the share of the canon a building's stream set can
  answer in principle, read from the KG without running any AI;
\item \emph{Realized readiness}: the share of the canon a fixed reference AI answers correctly,
  measured level by level, so that the rest of the canon splits into \emph{AI failure} (the data
  could answer, the AI did not), \emph{unprobed credit} (no question tested it), and \emph{data
  limitation} (the streams are absent), separating what a better system would fix from what only
  instrumenting would fix; and
\item \emph{Actuation-readiness}: the control handles a building exposes, so that a building's
  readiness is a pair of coverage measures, one for answering and one for acting.
\end{itemize}
The separation of data limitation from AI failure is the conceptual core of the paper. The ceiling
can be read from any KG, but measuring accuracy automatically and at scale requires the correct
answer to every question to be known in advance. The framework is therefore run on a published
simulated corpus of U.S.\ Department of Energy (DOE) prototype buildings, each a Brick KG linked to
physics-based time series, in which the answers follow from the simulation that generated the data,
fault labels are known, and instrumentation is varied as a controlled variable. It is then applied
to 45 public real-building KGs, 37 of which link their points to series, where no fault labels exist
and the data are as operators find them.

This study makes five contributions.
\begin{enumerate}
\item A measurement framework for building AI-readiness, built on these four constructs and
  computed from the KG and one fixed reference agent.
\item A question canon whose data requirements for diagnosis are derived from declared cause sets
  and checked against two practitioner corpora, with a question bank of 64 automatically graded
  classes that runs on simulated buildings and, where their data allow, on real ones.
\item A controlled study on a published simulated corpus with a pinned open-weight agent, repeated
  runs and archived transcripts, which measures how much of the ceiling the agent realizes at each
  instrumentation profile, and, on paired fault scenarios that look alike without the stream that
  tells their causes apart, whether the agent abstains when a cause is not identifiable and names
  the right one when it is.
\item A census of real buildings that reports both ceilings, the gap between what a KG declares
  and what its data support, the usability of the linked series, and how far the ranking of
  buildings depends on how the domains are weighted.
\item An open assessment tool: code, released under an open license, that computes both ceilings
  from any Brick KG, generates and grades the question bank, and runs the agent, together with the
  deposited results and the paired fault scenarios, so that a building owner can assess a building
  and a reader can regenerate every table and figure.
\end{enumerate}

\section{Related work}
\label{sec:related}

Three lines of prior work bear on measuring a building's AI-readiness: smart-readiness assessments
and building data standards, building QA over KGs, and fault diagnosis.

\subsection{Data readiness, smart readiness, and building data standards}
The notion that data carries a \emph{readiness} for a task predates its building-specific form.
Lawrence's data readiness levels~\cite{lawrence2017drl} propose a vocabulary for how prepared a
dataset is for analysis, by analogy to technology readiness levels. Within buildings, where
data-driven technologies increasingly mediate operation~\cite{billanes2025smartbuilding}, the
SRI~\cite{verbeke2020sri,ec2020sri} scores a building across nine technical domains and three
functionalities (energy performance and operation, occupant responsiveness, energy flexibility)
through an expert assessor cataloguing installed smart-ready services. It is, by design, a checklist
of capacity and not a measurement of realized capability, and its own methodology acknowledges that
a rating grounded in ``actual building performance data'' would be preferable but has not been
operationalized. When two expert panels assessed the same building with it, they had to reconcile
discrepancies between their assessments~\cite{vigna2020sri}. WiredScore's
SmartScore~\cite{wiredscore2026}, an international smart-building certification, likewise assesses
installed capability, from an expert scorecard in nine categories that run from digital services and
wellbeing to cybersecurity, data management and systems integration. \ref{app:sri} maps the SRI's 54
services to this study's questions.

On the data side, several conventions shape how a building's data is represented. Building
information models carry it from design into operation~\cite{parn2017bimfm}, with their content
graded by the level-of-development specification of BIMForum, a United States industry forum on
building information modelling~\cite{bimforum2024lod}, which describes how far each modelled element
is specified and not whether the element is sensed, and with recent work enriching these models with
real-time sensor streams for digital-twin monitoring~\cite{wang2022dtbim}. Brick and
ASHRAE~223P~\cite{ashrae223p,balaji2016brick,balaji2018brick}, with the Smart Applications REFerence
(SAREF) ontology family on the Internet of Things (IoT) side~\cite{saref2023etsi}, are
semantic-metadata schemas: a controlled vocabulary for describing a building's points, equipment,
and their relationships in a machine-readable KG. Building such KGs for real buildings is itself a
research problem, and the share of a building's points that end up modelled is a known measure of
metadata completeness~\cite{fierro2020shepherding,fierro2018mortar}. Application shapes, written in
the Shapes Constraint Language (SHACL) and supported by BuildingMOTIF~\cite{fierro2022buildingmotif}
and the 223P tooling, state what a KG must contain for a given application to run, and validate a KG
against that requirement. ASHRAE Guideline~36~\cite{ashrae_gdl36} is complementary, specifying
standardized sequences of operation for common HVAC systems together with the named input and output
points each sequence reads and writes. Such standards are a prerequisite for knowledge-graph
question answering (KGQA), since they make the KG exist and give its nodes a consistent meaning.
They govern how a building's data is described, however, not what a building measures. An
application shape checks the KG and not the series behind it, and a building can carry a complete,
valid KG and still lack the streams that a given operational question requires.

\subsection{Building question answering over knowledge graphs}
A closer line of evaluation work asks whether an AI can answer questions over a building's metadata
KG~\cite{jung2026bricktrace}, a building-specific case of the broader effort to pair LLMs with KGs
for QA~\cite{llmkg2025survey,avila2025kgqa}, with related building-facing chatbots over building
data~\cite{lin2023sitechatbot}. BuildingQA~\cite{mulayim2025buildingqa} contributes 188
question--answer pairs over four buildings (three in Brick, one in 223P), built from 27 practitioner
queries in SPARQL, the query language of KGs, with LLM-generated paraphrases, and reports that
zero-shot KGQA systems reach a top row-matching $F_1$ of only $0.38$, where row-matching $F_1$ is
the overlap between the result rows a generated query returns and the rows the ground-truth query
returns. Domain specific building KGQA is far from solved. Its ground-truth answers are the result
rows of the expert queries, so it grades whether an AI retrieves the right points and facts, such as
which air handler serves a zone, more than whether it computes a value from their series, such as
how many hours a zone was out of band on Tuesday. Computing over the series is the harder part,
because even strong LLMs struggle to reason zero-shot over raw time
series~\cite{merrill2024timeseries}. The Mortar testbed~\cite{fierro2018mortar} supplies the other
half: a library of analytics applications written by practitioners over Brick KGs and their series,
which this study uses as a second check on its question canon. Mortar also showed which of its
applications could run on which building by querying each KG for the points an application needs,
which amounts to a census of portability at the level of applications. The canon generalizes that
census from a list of applications to a declared set of questions, and the ceiling adds the check
that the series behind the points are usable.

\subsection{Fault diagnosis and diagnosability}
The fault-detection-and-diagnosis (FDD)
literature~\cite{katipamula2005fddreview,mirnaghi2020fddreview} contributes labelled fault
datasets~\cite{granderson2020data,granderson2023labeled} and an evaluation vocabulary: Frank et
al.~\cite{frank2019framework} define metrics for detection (true and false-alarm rates) and
diagnosis (correct-diagnosis rate), and digital-twin-enabled FDD continues this
line~\cite{xie2023dtfdd}. Classical FDD evaluates a diagnostic algorithm on a building whose
instrumentation is fixed, so it does not treat the availability of the stream that separates two
causes as a variable. Which faults a given set of sensors can tell apart is the subject of
diagnosability analysis and sensor placement in model-based
diagnosis~\cite{trave2006diagnosability,krysander2008sensorplacement}, which answers the question
from a model of the building. This study applies the same idea at a coarser level. It judges which
causes the data can tell apart from the classes of streams a building exposes and a declared list of
candidate causes, so the judgement can be read from any Brick KG without a model of the building,
and it becomes a property of the building that can be measured before any diagnostic algorithm is
chosen.

\subsection{Positioning of the present study}
Table~\ref{tab:positioning} sets the present study beside the constructs it is closest to. None of
them conditions its result on instrumentation or separates a data limitation from a model
limitation.

\begin{table}[!htbp]
\centering
\caption{The present study beside the constructs it is closest to.}
\label{tab:positioning}
\footnotesize\linespread{1}\selectfont
\setlength{\tabcolsep}{4pt}
\begin{tabular}{@{}>{\raggedright\arraybackslash}p{0.24\linewidth}>{\raggedright\arraybackslash}p{0.30\linewidth}>{\raggedright\arraybackslash}p{0.17\linewidth}>{\raggedright\arraybackslash}p{0.13\linewidth}>{\raggedright\arraybackslash}p{0.07\linewidth}@{}}
\toprule
Construct & What is scored & Evidence & \hspace{0pt}Conditioned on instrumentation & Data vs.\ AI \\
\midrule
SRI, SmartScore~\cite{verbeke2020sri,wiredscore2026} & installed smart services and their functionality level & assessor inspection & no & no \\
Data readiness levels~\cite{lawrence2017drl} & fitness of a dataset for analysis & data audit & no & no \\
Sensor coverage, metadata completeness~\cite{fierro2020shepherding} & points present or modelled & point counts & counts sensors, not answers & no \\
Application shapes~\cite{fierro2022buildingmotif} & whether a KG holds what an application needs & KG validation & KG only, not the series & no \\
Building KGQA benchmarks~\cite{mulayim2025buildingqa,jung2026bricktrace} & an AI's retrieval from the KG & ground-truth queries & no & no \\
This study & the share of a question canon the data can answer, and the share a fixed agent realizes & KG, series, and a fixed agent & yes & yes \\
\bottomrule
\multicolumn{5}{@{}p{0.98\linewidth}@{}}{\footnotesize ``Data vs.\ AI'': whether the construct separates a limitation of the building's data from a limitation of the AI.}
\end{tabular}
\end{table}

\FloatBarrier

\section{Building AI-Readiness Assessment Framework}
\label{sec:framework}

The framework holds the AI constant across buildings, so that a difference in the readiness measures
reflects a difference between buildings. The premises of the framework are: (1) a question is
considered answerable only when the building's data can determine the answer; (2) a domain is
considered actuable only when the building's KG exposes a handle for the action (a setpoint or a
command); and (3) the building's KG is assumed to be complete and correct, meaning that it models
every point the building has and that every point it declares exists and records data
(Section~\ref{sec:kg-quality} discusses what happens when this assumption fails).

\begin{figure}[tbp]
\centering
\makebox[\linewidth][c]{\includegraphics[width=1.2\linewidth]{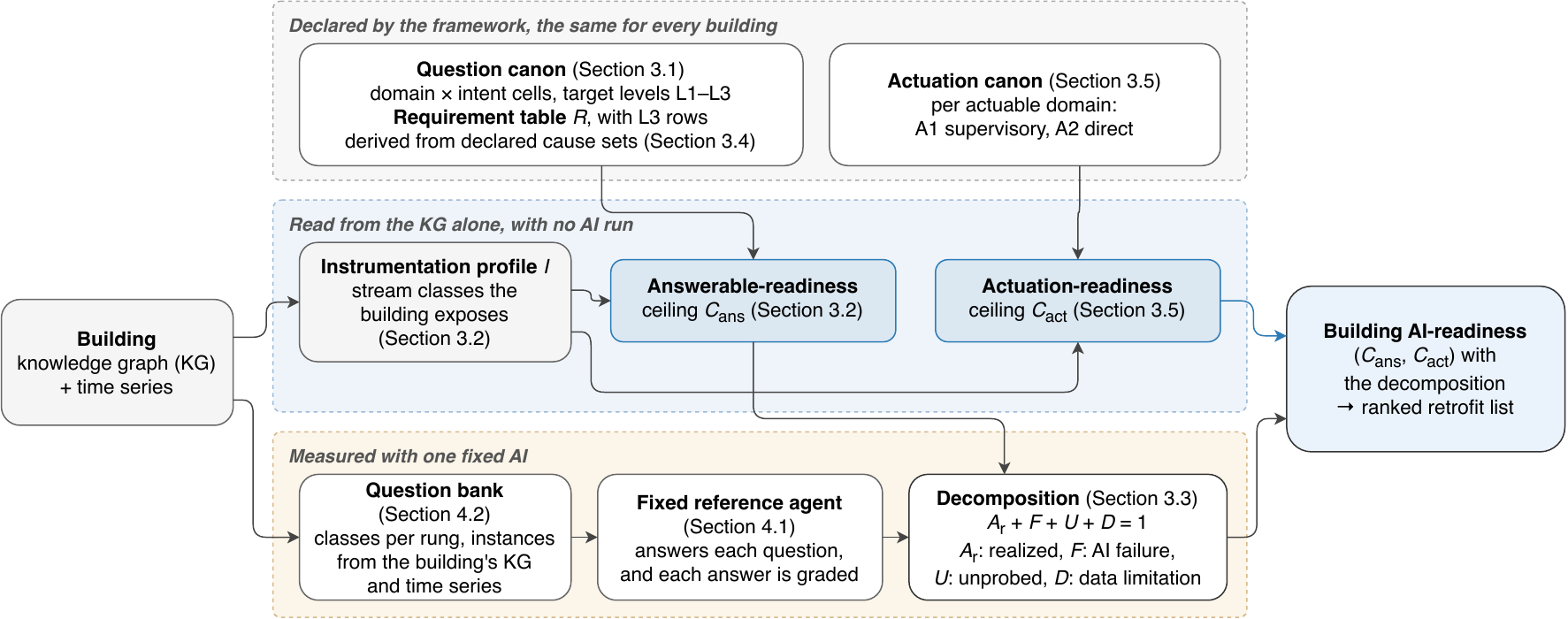}}
\caption{Overview of the framework. The section numbers show where each step is defined.}
\label{fig:overview}
\end{figure}

Figure~\ref{fig:overview} shows how the constructs connect, and Table~\ref{tab:nomenclature} lists
the terms and symbols that recur in later sections.

\begin{table}[tbp]
\centering
\caption{Key terms and symbols.}
\label{tab:nomenclature}
\footnotesize\linespread{1}\selectfont\renewcommand{\arraystretch}{0.95}
\begin{tabular}{@{}>{\raggedright\arraybackslash}p{0.27\linewidth}
  >{\raggedright\arraybackslash}p{0.10\linewidth}>{\raggedright\arraybackslash}p{0.56\linewidth}@{}}
\toprule
Term & Symbol & Meaning \\
\midrule
Canon & & the declared set of operational questions, as domain--intent cells with target levels
  (question canon), and the declared control rungs of each actuable domain (actuation canon) \\
Cell & $(d,i)$ & one domain--intent pair of the canon (Table~\ref{tab:canon}) \\
Level, target level & L1--L3 & answer completeness, from L1 reporting to L2 judging to L3
  attributing; the target level is the highest level a cell asks for \\
Rung, canon total & $\cS$ & one level of one cell; the canon has 76 rungs \\
Stream class & & a kind of measured or commanded point, defined as a set of Brick classes \\
Instrumentation profile & $I$ & the set of stream classes a building exposes \\
Requirement table, achieved level & $R$, $\ell_{d,i}(I)$ & the stream classes each level of a cell
  needs; the highest level whose classes are all in $I$ \\
Candidate causes, separating stream & & the declared causes of a deviation in a domain, and the
  stream class that tells them apart \\
Answerable-readiness ceiling & $\CI$ & the share of the canon that the data can support \\
Handle, actuation rung & A1, A2 & a point through which an agent can act on a domain: one that sets a target or a mode (A1, supervisory) or one that drives the device (A2, direct) \\
Actuation-readiness ceiling & $\Cact$ & the share of the actuation rungs that the KG's handles
  reach \\
Reference agent & & the fixed AI, one local model with three tools, that answers the questions in
  every building \\
Question class, question bank & & a template that poses questions for one rung and computes their
  ground truth; the bank is the set of 64 classes, and a rung is probed when the bank poses
  questions for it \\
Episode, turn & & one attempt of the agent at one question; one reply of the model within it \\
Accuracy & $\Acc$ & the reference agent's mean accuracy over the probed rungs \\
Realized readiness, AI failure, unprobed credit, data limitation & $\Areal$, $\aifail$, $\unp$,
  $\dlim$ & the four shares of the canon, which sum to one: answered correctly, answered wrongly,
  credited but not tested, and not supported by the data \\
Overclaim rate & & the share of unanswerable questions that the agent answered \\
Declared, linked, usable & & a point described in the KG; a point with a time series; a point
  whose series passes the usability rule \\
\bottomrule
\end{tabular}
\end{table}

\subsection{The question canon}
\label{sec:canon}
This study organizes the operational questions asked of a building along three axes. A question's
\emph{domain} is the physical subsystem it concerns: thermal, energy, air quality, equipment,
occupancy, or lighting. These are the technical domains of the SRI~\cite{verbeke2020sri} and the
subsystems that Brick organizes~\cite{balaji2018brick}. Plug loads and thermal energy are end uses
inside the energy domain, and illuminance is a lighting quantity. Acoustics is left out because a
building automation system records no stream from which an acoustic question has a verifiable
answer. A question's \emph{intent} is the operation requested: \emph{status} (the current value),
\emph{summary} (an aggregate over time or space), \emph{comparison} (against a peer, a baseline, or
a prior period), \emph{verification} (is the intended behaviour occurring), \emph{diagnosis} (which
or where is the problem), or \emph{explanation} (why). The intents form a progression of increasing
inferential demand, from reporting a measured value to explaining a cause. Predictive and
prescriptive intents are excluded because the building's data do not determine their answers. A
forecast can be graded against held-out data, but its accuracy measures the forecasting model more
than the building's data. A recommendation is an action, and its quality has no automatically
gradable answer. Grading what the data determine is what makes the framework a measurement. A
request to change the building, such as lowering a setpoint, belongs to the actuation axis
(Section~\ref{sec:actuation}).

Table~\ref{tab:canon} lists the canon as a matrix of domains and intents. Each entry is a
\emph{cell}, and its value is the cell's \emph{target level}, the highest level on a three-step
ladder of answer completeness that its questions ask for. At \textbf{L1} (\emph{reporting}) the
answer reports the quantity as the building automation system records it, such as a zone temperature
or a meter reading. At \textbf{L2} (\emph{judging}) the answer also judges that quantity against a
reference, such as a setpoint, the zone's own history, a peer zone, or a rated capacity. At
\textbf{L3} (\emph{attributing}) the answer also attributes a deviation to one cause among a
declared set of candidate causes (Section~\ref{sec:identifiability}). For the diagnosis and
explanation intents, the lower levels prepare that step. L1 locates or sizes the deviation, such as
the warmest zone or the change between two days, and L2 judges it against a reference. L3 then names
the cause from the domain's declared set, such as a stuck damper or a changed schedule for a warm
zone, by reading the stream that separates that cause from the others.

The intent and the domain set the target level of a cell. Status and summary questions report or
aggregate a quantity, so they are complete at L2. Energy status is complete at L1, because a meter
reading is reported without a reference. Comparison, verification, diagnosis and explanation
questions target L3 in the domains that declare a set of candidate causes, which are thermal,
energy, air quality and equipment. Occupancy and lighting declare none, because occupancy is the
occupants' choice and lighting follows occupancy or a schedule, so their cells stop at L2. L3 is the
last level, and tracing a named cause to a deeper root cause is left to future work. The
instrumentation sets the level a building achieves. One level of one cell is a \emph{rung}, so a
cell whose target level is L3 has three rungs. The canon has $\cS=76$ rungs, its \emph{canon total}.
A cell is left out where another cell already covers its question or where the intent does not
apply. Energy has no verification cell, because checking consumption against its expected value is
the comparison intent. Occupancy has none, because no intended behaviour is set for occupants, and
lighting has none, because checking that lights follow occupancy is lighting status at L2. Without
candidate causes, explaining a change in occupancy or lighting is the same question as diagnosing
it, so neither domain has an explanation cell.

\begin{table}[tbp]
\centering
\caption{The question canon: the target level (1--3) of each (domain, intent) cell. ``--'' marks a
cell left out of the canon, for the reasons given in the text. The target levels sum to the canon
total $\cS=76$.}
\label{tab:canon}
\footnotesize
\setlength{\tabcolsep}{4pt}
\begin{tabular}{@{}lcccccc@{}}
\toprule
\textbf{Domain} & Status & Summary & Comparison & Verification & Diagnosis & Explanation \\
\midrule
Thermal     & 2 & 2 & 3 & 3 & 3 & 3 \\
Energy      & 1 & 2 & 3 & -- & 3 & 3 \\
Air quality & 2 & 2 & 3 & 3 & 3 & 3 \\
Equipment   & 2 & 2 & 3 & 3 & 3 & 3 \\
Occupancy   & 2 & 2 & 2 & -- & 2 & -- \\
Lighting    & 2 & 2 & 2 & -- & 2 & -- \\
\bottomrule
\end{tabular}
\end{table}

Each level of a cell requires a set of \emph{stream classes}. A stream class is a kind of measured
or commanded point, defined as a set of Brick classes so that it can be read from any Brick KG. A
building exposes a stream class when at least one point of that class carries a usable time series.
A value stored in the KG itself, such as a setpoint's design value, does not expose a stream class.
Requirements are cumulative. Thermal status needs a zone temperature sensor at L1 and a zone
temperature setpoint, heating or cooling, at L2. Thermal verification also needs, at L3, the
terminal unit's actuation command and its response, such as a damper command and the damper position
or the airflow it produces. A peer zone's temperature is the same stream class on another zone, so a
comparison with a peer adds no requirement. Air quality is read through zone CO$_2$ alone. CO$_2$
tracks the occupant-generated load and the outdoor-air supply, which are the two causes that the
air-quality cells attribute, and it is the quantity that demand-controlled ventilation acts on.
Particulate matter, volatile organic compounds and humidity have other causes and are outside this
canon. Table~\ref{tab:canon-req} gives the full requirement table.

\begin{table}[tbp]
\centering
\caption{The requirement table $R$: the stream classes a cell needs at each level. Each entry lists
every class needed to reach that level. ``--'' marks a level above the cell's target level.}
\label{tab:canon-req}
\footnotesize\linespread{1}\selectfont
\setlength{\tabcolsep}{4pt}
\begin{tabular}{@{}l >{\raggedright\arraybackslash}p{0.26\linewidth} p{0.09\linewidth} p{0.14\linewidth} p{0.29\linewidth}@{}}
\toprule
Domain & Intent(s) & L1 & L2 & L3 \\
\midrule
Thermal & status, summary & T & T+Sp & -- \\
Thermal & comparison & T & T & T+Sp+ActC+ActP \\
Thermal & verification & T & T+Sp & T+Sp+ActC+ActP \\
Thermal & diagnosis, explanation & T & T+Sp & T+Sp+SpC+\allowbreak ActC+ActP+SAT \\
Energy & status & Whole & -- & -- \\
Energy & summary & Whole & Whole+Sub & -- \\
Energy & comparison, diagnosis, explanation & Whole & Whole & Whole+Sub2+OAT \\
Air quality & status, summary & CO2 & CO2+Occ & -- \\
Air quality & comparison & CO2 & CO2 & CO2+Occ+OAF \\
Air quality & verification, diagnosis, explanation & CO2 & CO2+Occ & CO2+Occ+OAF \\
Equipment & status, summary & Pwr & Pwr & -- \\
Equipment & comparison, verification, diagnosis, explanation & Pwr & Pwr & Pwr+Run+Load \\
Occupancy & status, summary, comparison, diagnosis & Occ & Occ & -- \\
Lighting & status, diagnosis & Lgt & Lgt+Occ & -- \\
Lighting & summary, comparison & Lgt & Lgt & -- \\
\midrule
\multicolumn{5}{@{}>{\raggedright\arraybackslash}p{0.95\linewidth}@{}}{\footnotesize Stream classes: T~=~zone air temperature;
Sp~=~zone temperature setpoint in effect, heating or cooling; SpC~=~scheduled (occupied or
unoccupied) setpoint held apart from the setpoint in effect; ActC~=~terminal-unit actuation command
(damper, valve or reheat command); ActP~=~terminal-unit actuation response (damper or valve
position, or terminal airflow); SAT~=~supply air temperature of the
air system serving the zone; OAT~=~outside air temperature; Occ~=~occupancy (presence or count);
CO2~=~zone CO$_2$ concentration; OAF~=~outdoor-air flow; Whole~=~whole-building electricity meter;
Sub~=~at least one end-use submeter (HVAC, lighting, plug loads or thermal energy); Sub2~=~two
end-use groups metered, which fixes the third as the residual; Pwr~=~equipment input power
(electricity or fuel); Run~=~equipment run command or status; Load~=~equipment load (air or water
flow, or delivered heating or cooling); Lgt~=~lighting power, lighting status, or illuminance.} \\
\bottomrule
\end{tabular}
\end{table}

The canon is declared in full so that it can be audited and replaced. Besides Tables~\ref{tab:canon}
and~\ref{tab:canon-req}, \ref{app:canon} gives an example question at each level of every cell, the
candidate causes that each domain declares, and the Brick classes that define each stream class. The
canon and the question bank were written for this study and checked against two practitioner
corpora. Of the 27 expert queries in BuildingQA~\cite{mulayim2025buildingqa}, 21 look up a stream
class the canon names, and the others ask for every point of an entity or for metadata. Every
quantity computed by the analytics applications published with the Mortar
testbed~\cite{fierro2018mortar} falls in a canon cell, and four of the applications read the supply
air temperature or the outside air temperature, which is why the canon includes both streams. All
scores are relative to this canon, and a site can substitute its own questions or weight the domains
without changing the method. Domain experts enter at three points. They audit the canon's cells,
target levels and cause sets for a building type, they choose a site's domain weights, and they
judge whether a declared point can be trusted before its series is used. Reweighting the domains can
reorder two buildings whose strengths lie in different domains, but never the nested profiles of one
building, because adding a stream class can only raise the level of a cell.

\subsection{The answerable-readiness ceiling}
\label{sec:ceiling}
\begin{figure}[h]
\centering
\includegraphics[width=\linewidth]{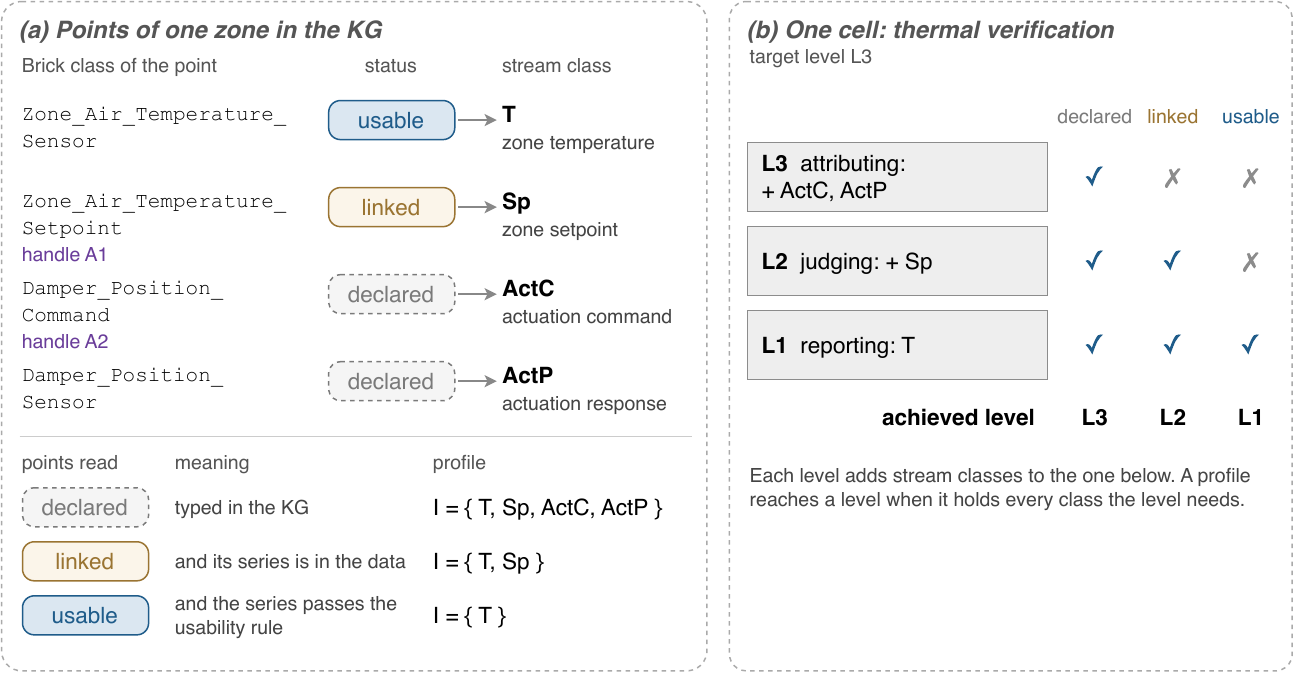}
\caption{From a zone's points to a cell's achieved level. (a) The usability rule
(Section~\ref{sec:realq}) drops series with large gaps or stuck values. Each point maps to a stream
class (Table~\ref{tab:brickmap}), and the stream classes of the points read form the instrumentation
profile $I$, so one zone has three nested profiles. A setpoint or command point is also a handle, A1
or A2, for the actuation-readiness ceiling (Section~\ref{sec:actuation}). (b) The cell reaches the
highest level whose required stream classes are all in the profile: L3 on the declared points, L2 on
the linked points and L1 on the usable points.}
\label{fig:ladder}
\end{figure}

The \emph{instrumentation profile} $I$ of a building is the set of stream classes it exposes. The
\emph{requirement table} $R$ (Table~\ref{tab:canon-req}) lists, for the cell of domain $d$ and
intent $i$, the stream classes $R_{d,i,r}$ needed to reach each level $r$ of the ladder, up to the
cell's target level $c_{d,i}$. The \emph{achieved level} $\ell_{d,i}(I)$ is the highest level whose
required classes are all in $I$ (Figure~\ref{fig:ladder}),
\begin{equation}
\label{eq:level}
  \ell_{d,i}(I) \;=\; \max\,\{\,r : R_{d,i,r} \subseteq I \,\}, \qquad \ell_{d,i}=0 \text{ if even L1 is unmet.}
\end{equation}
The \emph{answerable-readiness ceiling} $\CI$ is the share of the canon total that the profile
achieves. The subscript marks the answering axis, and the argument marks the dependence on $I$,
\begin{equation}
\label{eq:ci}
  \CI \;=\; \frac{1}{\cS}\sum_{d,i}\ell_{d,i}(I)\ \in[0,1].
\end{equation}

$\CI$ depends only on the stream classes the building exposes, not on the AI, the data values or the
wording of a question, so it is the most that a perfect AI could realize. It rises in steps as
stream classes are added, either by installing a sensor or meter and entering it in the KG, or by
mapping a point the automation system already has into the KG and linking its series, which needs no
new hardware.

The ceiling counts every rung equally, and a site that values some domains more can weight them.
With domain weights $w_d$, the weighted ceiling is
\begin{equation}
\label{eq:cw}
  C_{w}(I) \;=\; \frac{\sum_{d} w_d \sum_{i}\ell_{d,i}(I)}{\sum_{d} w_d \sum_{i} c_{d,i}},
\end{equation}
which returns $\CI$ when all weights are equal, because the target levels $c_{d,i}$ sum to $\cS$.
Counting each domain equally is a different choice. It sets $w_d$ to $1/\sum_i c_{d,i}$, so that
every domain's own share enters with the same weight.

\subsection{Realized readiness and the decomposition}
\label{sec:decomp}
\begin{figure}[tbp]
\centering
\includegraphics[width=0.82\linewidth]{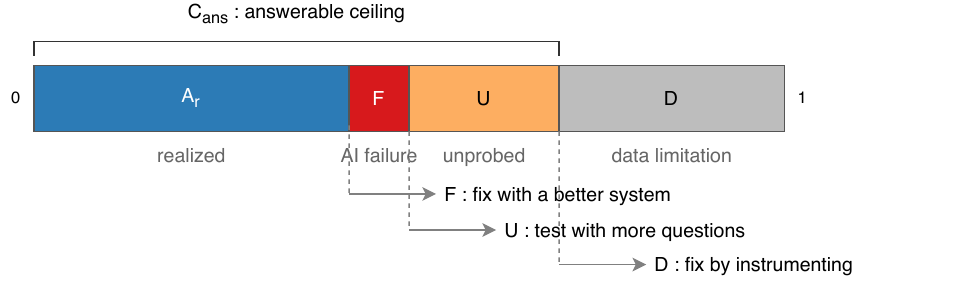}
\caption{Conceptual decomposition of building AI-readiness. The canon total splits into realized
readiness, AI failure, unprobed credit and data limitation.}
\label{fig:decomp-concept}
\end{figure}

A building realizes only part of its ceiling. The ceiling credits one rung for every achieved level
of every cell, so $\cS\,\CI$ rungs are credited. Each question class targets one rung, and the
reference agent's accuracy is measured rung by rung. Let $a_{d,i,r}$ be the share of the questions
targeting level $r$ of cell $(d,i)$ that the agent answers correctly, and let $P(I)$ be the set of
credited rungs that the question bank probes, and $|P(I)|$ their number. A credited rung outside
$P(I)$ is \emph{unprobed}, and its accuracy is not assumed. The \emph{realized readiness} is the
share of the canon that is credited, probed and answered correctly,
\begin{equation}
\label{eq:areal}
  \Areal \;=\; \frac{1}{\cS}\sum_{(d,i,r)\in P(I)} a_{d,i,r}.
\end{equation}
The whole canon then splits into four shares (Figure~\ref{fig:decomp-concept}),
\begin{equation}
\label{eq:decomp}
\begin{split}
  1 \;=\; \Areal
  \;&+\; \underbrace{\frac{1}{\cS}\sum_{(d,i,r)\in P(I)}\bigl(1-a_{d,i,r}\bigr)}_{\textstyle
    \aifail:\ \text{AI failure}}\\[1.5ex]
  \;&+\; \underbrace{\CI-\frac{|P(I)|}{\cS}}_{\textstyle \unp:\ \text{unprobed}}
  \;+\; \underbrace{\bigl(1-\CI\bigr)}_{\textstyle \dlim:\ \text{data limitation}} .
\end{split}
\end{equation}
The \emph{data-limitation} share $\dlim$ is readiness lost because streams are absent, and only
instrumenting recovers it. The \emph{AI-failure} share $\aifail$ is readiness lost on questions the
data could answer, and a better agent recovers it. The \emph{unprobed} share $\unp$ is credit the
measurement has not tested, and a larger question bank shrinks it. When the question bank probes
every credited rung, $\unp=0$ and $\Areal=\CI\,\Acc$, where $\Acc$ is the mean of $a_{d,i,r}$ over
the probed rungs. When it does not, the product $\CI\,\Acc$ credits the unprobed rungs with the
accuracy of the probed ones. The four shares are therefore always reported together, on the canon
total.

Questions are generated from the building's own KG, so a question about equipment the building does
not have is never posed. Such a question is \emph{not applicable} and counts nowhere. Applicability
acts on questions, not on cells. The cells of the canon are defined over the six service domains,
which every occupied building has, so the ceiling keeps the full canon total as its denominator, and
a building without the data for a domain is scored as data-limited in it.

A generated question is \emph{answerable} under a profile when the streams its level requires are
exposed. On an answerable question the agent is either correct or it is an AI failure, which covers
a wrong answer and an abstention the data did not force. On an unanswerable question an abstention
is correct, and any answer is an \emph{overclaim}, even when the answer happens to be right, because
the data did not determine it. Only answerable questions enter $a_{d,i,r}$. The \emph{overclaim
rate} is the share of unanswerable questions that the agent answered, and it measures whether the
agent respects the ceiling. Table~\ref{tab:outcomes} summarizes the four outcomes.

\begin{table}[tbp]
\centering
\caption{How each answer is scored. Answerable questions enter the rung accuracy $a_{d,i,r}$, and
unanswerable ones enter the overclaim rate.}
\label{tab:outcomes}
\small\linespread{1}\selectfont
\begin{tabular}{@{}lll@{}}
\toprule
 & Agent answers & Agent abstains \\
\midrule
Answerable question & correct if right, AI failure if wrong & AI failure \\
\addlinespace
Unanswerable question & overclaim, even if right & correct abstention \\
\bottomrule
\end{tabular}
\end{table}

\subsection{Identifiability of causes}
\label{sec:identifiability}
A cell is answerable at a level only if the available streams \emph{uniquely determine} the answer.
This matters little at L1 and L2, where a value and its reference settle the answer, and is decisive
at L3. A cause is identifiable only when the data separate it from every other cause that would
produce the same observations, so identifiability is defined against a declared set of candidate
causes. For each domain with L3 cells, the canon declares a small cause set drawn from the fault
detection and diagnosis literature~\cite{katipamula2005fddreview,frank2019framework}. For each cause
it names the stream class whose behaviour sets that cause apart from its rivals
(Table~\ref{tab:causes} in \ref{app:canon}). The L3 requirement of a diagnosis or explanation cell
is the L2 requirement plus the union of the separating classes of its domain. A comparison or
verification cell at L3 attributes one observed difference or deviation, and it needs the separating
classes of the causes that difference can have. In the thermal domain a difference between zones, or
a deviation from the setpoint, has three kinds of cause. The zones may be asked for different
temperatures, which the setpoint shows. The terminal may be stuck, which shows as a command that
moves while its response stays. Or the zone may carry a different load, which shows as command and
response moving together. Thermal comparison and verification at L3 therefore need the setpoint, the
terminal command and its response. In the other domains the comparison and verification cells that
reach L3 share the diagnosis requirement. A profile that holds a related stream and still cannot
separate two declared causes does not reach L3. Equipment power alone, for example, shows that a
unit's consumption rose, and only its run status and load tell a commanded change from a load change
and from degradation.

\subsection{The second axis: actuation-readiness}
\label{sec:actuation}
Answering is only half of autonomy. An agent that can report a zone's state but cannot drive any
actuator is not ready to operate the building. Control handles, like sensors, are part of the
instrumentation and can be read from the same KG. This study therefore defines a second axis, the
\emph{actuation-readiness ceiling} $\Cact$, built like the answerable-readiness ceiling but over
control handles. It records whether the building exposes the handle needed to act on each domain.
How well an agent would control is not measured, because a control policy has no automatically
gradable ground truth, so actuation enters as a ceiling only and no agent is run on it.

\begin{table}[!tb]
\centering
\caption{The actuation canon: the Brick classes that make up each rung of each actuable domain.}
\label{tab:actcanon}
\small
\begin{tabular}{@{}lp{0.34\linewidth}p{0.40\linewidth}@{}}
\toprule
Domain & A1 (supervisory) & A2 (direct) \\
\midrule
Thermal & zone cooling/heating setpoint & damper / valve / reheat command \\
Air quality & airflow / fan-speed setpoint & outside-air damper command \\
Equipment & on/off, enable or start/stop command & compressor command, speed setpoint \\
Lighting & lighting-level setpoint & lighting on/off / luminance command \\
\bottomrule
\end{tabular}
\end{table}

Four of the six domains can in principle be acted on in any building: thermal, air quality through
ventilation, equipment, and lighting. Energy can be acted on only where the building has
dispatchable electrical assets, such as a battery, a photovoltaic inverter or managed vehicle
charging, whose power setpoints an agent can write. Every building can meter its energy, so the
answering axis keeps energy in its denominator, while the actuation axis leaves energy out where
there are no such assets, as in every building studied here. Occupancy is never actuable. Each
actuable domain is scored on a two-rung ladder parallel to the L1/L2/L3 answer ladder. At
\textbf{A1} (\emph{supervisory}) the KG exposes a point that sets a target or a mode, which the
domain's local controller then carries out, such as a zone setpoint or an equipment enable command.
At \textbf{A2} (\emph{direct}) it also exposes a point that drives the device itself, such as a
damper position, a valve or a compressor command. The achieved rung is cumulative (A2 requires A1),
and
\begin{equation}
\label{eq:cact}
  \Cact \;=\; \frac{\sum_{d\in\mathcal{D}_{\mathrm{act}}}a_d(I)}{2\,|\mathcal{D}_{\mathrm{act}}|}\ \in[0,1],
\end{equation}
where $a_d(I)\in\{0,1,2\}$ is the achieved actuation rung of domain $d$ and
$\mathcal{D}_{\mathrm{act}}$ is the set of actuable domains, four in every building studied here, so
the denominator is 8. Table~\ref{tab:actcanon} lists the Brick classes that make up each rung, which
form the \emph{actuation canon}. The same zone setpoint can appear in both canons, as a reference
read to judge a temperature and as a handle written to change the target.

\begin{figure}[!tb]
\centering
\includegraphics[width=0.45\linewidth]{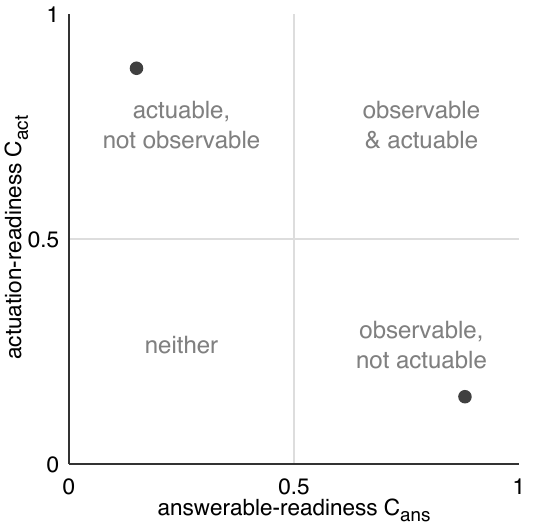}
\caption{AI-readiness of a building as a point in the plane of the two ceilings. The two points are
illustrative.}
\label{fig:axes}
\end{figure}

\paragraph{Two axes} A building's readiness is thus a pair, answerable-readiness and
actuation-readiness (Figure~\ref{fig:axes}). A building can be observable but not actuable, or
actuable but not observable. An agent could then be asked to diagnose a fault it cannot fix, or to
drive a system it cannot see. The two axes are therefore reported side by side and never collapsed
into a single index, which would hide this asymmetry. One caveat weighs more on the actuation axis.
A setpoint or command point in the KG shows that the handle is modelled and addressable. Whether the
deployed automation system lets an external agent write to it is an integration property that no KG
records, so $\Cact$ is an upper bound on what an agent may drive.

\FloatBarrier

\section{Measurement Substrate and Study Design}
\label{sec:methods}

\begin{figure}[!htbp]
\centering
\includegraphics[width=0.95\linewidth]{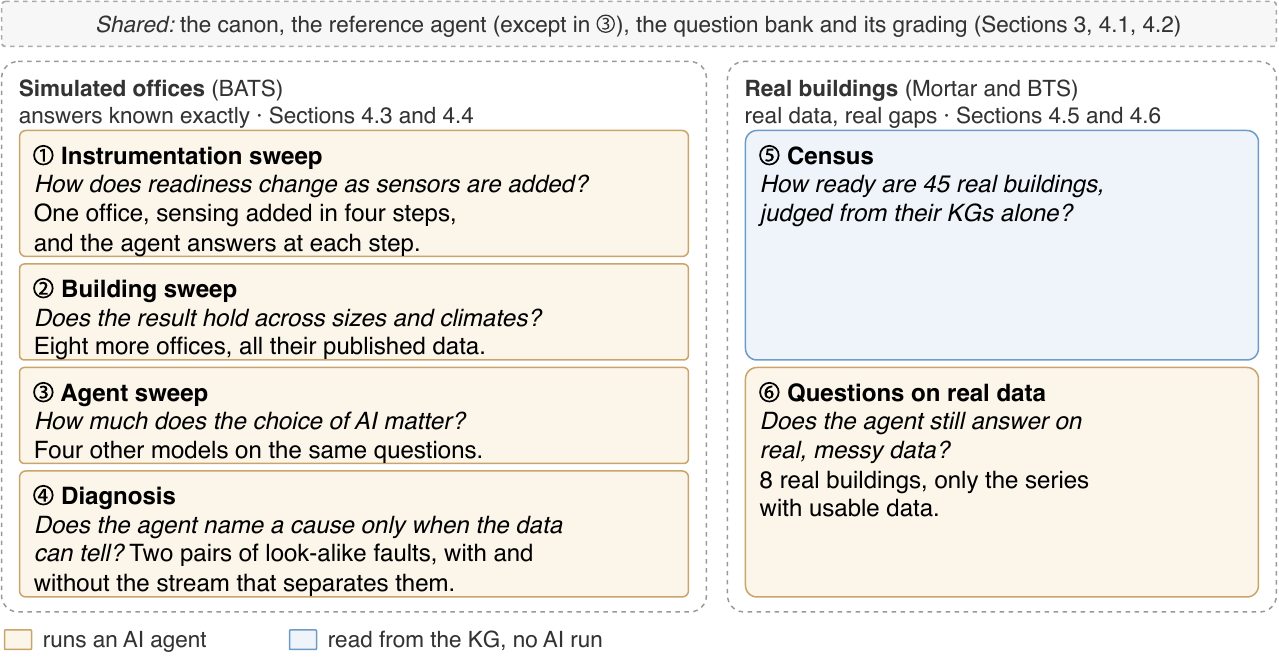}
\caption{Study design: the question each experiment answers and the data it uses. The top band and
the column headings give the sections that describe them.}
\label{fig:design}
\end{figure}

The study rests on two evidence bases. The first is a controlled study on a published simulated
corpus, in which instrumentation is varied as the only independent variable, the true cause of every
fault and the true value behind every reading are known, and the reference agent is probed with the
answer withheld. The second is a census of real buildings, in which the two ceilings are read from
real Brick KGs and part of the question bank is run on real series. The simulated data are complete
and clean, and the real buildings show what simulation cannot: which points real KGs actually link,
and how many of their series are usable. Figure~\ref{fig:design} lists the experiments, the question
each one answers, and the data it uses.

\subsection{The reference agent}
\label{sec:agent}
The system under test is a fixed agent built on an open-weight language model, Qwen3.8 with 27
billion parameters~\cite{qwen38}, run locally in a compressed four-bit form. An open-weight model is
used for three reasons. Its weights can be reloaded after a hosted model is retired, so a readiness
score stays reproducible. It runs locally, so a building's data stay on site. And it costs nothing
per question, so an assessment can be repeated as the building changes. The reference agent need not
be the strongest model available, because the measurement concerns the building. A stronger agent
can only shrink the AI-failure share, which at L1 and L2 stays small next to the ceiling
(Section~\ref{sec:res-sweep}).

\begin{figure}[!htb]
\centering
\includegraphics[width=0.90\linewidth]{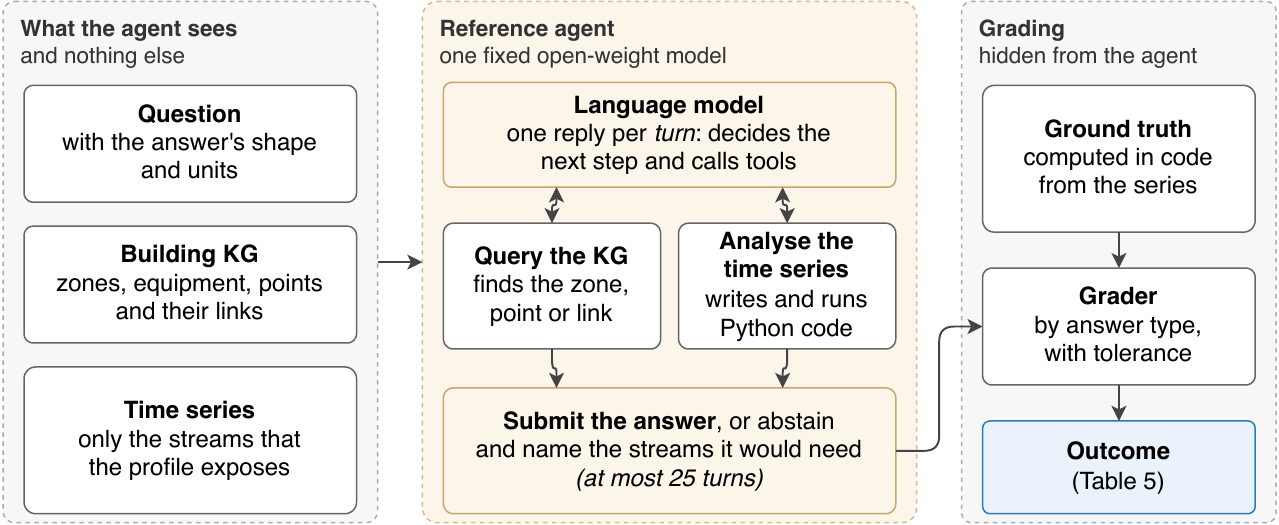}
\caption{One question episode. A \emph{turn} is one reply of the model, with the tools it calls and
their results. An episode ends when the agent submits, when it replies without calling a tool, or
after 25 turns. Table~\ref{tab:episode} follows one episode call by call.}
\label{fig:agent}
\end{figure}

The agent has three tools (Figure~\ref{fig:agent}). A KG query tool runs a SPARQL query over the
building's KG, a Python tool computes over the series in an isolated process, and an answer tool
hands in the answer in the shape the question names, or abstains and names the streams the agent
would need. The model decides when to call each tool. There is no fallback model and no retry after
a wrong answer, and an episode ends with no answer after 25 turns. Sampling is left at the model
server's defaults, so repeated runs measure how much the answers vary. \ref{app:agent} gives the
model build, the prompt, the tool schemas and the token counts, and the agent's code is released
with the study's code.

Two choices keep accuracy a fair measurement. First, the agent sees only the question, the KG and
the series of the active profile, never the ground-truth answer or the code that computes it.
Second, a stream the profile lacks is absent from those series, so a question that needs it is
unanswerable to the agent. Each question also states its answer shape and units, so grading targets
retrieval and reasoning. These choices guard against the shortcuts a model can exploit in a
carelessly designed benchmark~\cite{geirhos2020shortcut}.

\subsection{The question bank and its grading}
\label{sec:bank}
The canon of Section~\ref{sec:canon} defines the question space. The question bank has 64 question
classes over it (\ref{app:bank}), each bound to one cell at one level. Together they target 59 of
the 76 rungs, among them every rung that the published corpus credits at any profile. Five classes
reproduce analyses from the Mortar application library, and four span two KG hops, such as the
warmest zone under a named air handler. A class requires at least the streams that the canon
requires of its rung, so a question is posed only on a rung that the ceiling credits. The L3 classes
compute the quantity that separates the causes, such as the damper signal behind a temperature
difference, and grade that quantity. They test whether an agent can reach and use the separating
stream that attribution needs, and the diagnosis scenarios of Section~\ref{sec:faults} test the
attribution itself. An L3 rung's accuracy is therefore an upper bound on how well an agent
attributes.

Instances are sampled from the building's own KG and series, up to three per class from a fixed
seed, over zones, equipment, dates and seasons, and the ground-truth answer is computed in code from
the same series. A class whose streams the building lacks yields no instances, and its rungs count
as data limitation. Every question states its own conventions, such as the window, the units and the
sign of a difference, so that a disagreement over conventions is not scored as an AI failure.

Each answer is scored by a grader matched to its type, fixed in advance. A \emph{numeric} answer is
correct within a tolerance set per class, such as $\pm0.5$\,\si{\celsius} for a temperature,
following benchmarks for numerical question answering~\cite{chen2021finqa,zhu2021tatqa}. A
\emph{boolean} answer must match the ground-truth verdict, and an \emph{entity} answer must name the
ground-truth entity by any name the KG ties to it. Every episode is then classed by the outcomes of
Table~\ref{tab:outcomes}. On the sweep building each question is asked three times, on the other
buildings once, and the accuracy of a rung is the mean over its instances and repeats. Realized
readiness is reported with a 95\% bootstrap interval, every other proportion with its Wilson 95\%
interval, and a difference between two small proportions is tested with Fisher's exact test
(\ref{app:bank}). A guessing baseline, a fixed answer chosen in advance for each answer type, is
scored on the same questions.

\subsection{The simulated corpus}
\label{sec:substrate}
The corpus is the Building AI Testbed Suite (BATS)~\cite{bats2026}, a published dataset generated
with the BuildStream simulation software from DOE commercial prototype models~\cite{doe_prototypes}
simulated in EnergyPlus~\cite{crawley2001energyplus}. It holds nine clean office buildings, the
small, medium and large office prototypes in three climates (Tampa, 2A; New York, 4A; Buffalo, 5A),
all at the ASHRAE~90.1-2019 vintage. Each has a faulted variant with labelled faults. Every building
comes as a Brick~\cite{balaji2018brick} KG and hourly time series for one year. Each point that has
a series carries a reference id that resolves to its series, so an agent can go from a KG entity to
its measurements. A high accuracy cannot come from answers the model memorized in training, because
the correct answer differs by building and climate and a numeric answer has to be computed from the
series. Template-generated benchmarks with controlled ground truth are an established design in
machine learning~\cite{johnson2017clevr}.

Three features of the published data affect the study, and each is handled before any question is
asked. First, the KG describes more points than the data include. In the large office, 761 points
are described as having a series, and the release contains 93 of them. The ceiling is therefore
computed both on every point the KG describes and on the points with data, and the gap between the
two shows how far the corpus breaks the third premise of Section~\ref{sec:framework}. Second, the KG
records equipment links from the equipment's side and zone links from the point's side. The study
adds every such link in the other direction, a routine step with KGs, and runs four question classes
on the original KG to test whether this matters. Third, the zone-level power sensors carry the
simulated zone lighting power under a generic type. The study counts zone-attached power as
lighting, in the ceiling and in the text of every lighting question, so that the ceiling and the
agent apply the same rule.

\subsection{Experiments on the simulated corpus}
\label{sec:sim-exp}
\paragraph{Instrumentation sweep}\label{sec:sweep}
The sweep varies instrumentation on one building, the large office in Tampa, and holds everything
else fixed. A profile keeps only the series of its stream classes, which models ``this building
lacks that sensor'' while the physics, the building and the climate stay fixed. Five nested profiles
are used (Table~\ref{tab:profiles}). The corpus buildings are at the 90.1-2019 vintage, and the
metered profile takes its metering from the current edition,
ASHRAE~90.1-2022~\cite{ashrae90_1_2022}, whose Section~8.4.3 requires the total electrical energy
and the HVAC, interior lighting, exterior lighting, receptacle and refrigeration energy to be
metered separately. The total is the whole-building meter and the end uses are the submeters of
Table~\ref{tab:canon-req}. The rich profile adds the points that the Guideline~36
sequences~\cite{ashrae_gdl36} read and write. On the published corpus the metered and rich profiles
coincide, because the release has none of the rich profile's points, so the sweep runs four distinct
profiles and the rich profile's points appear only in the real buildings. For each profile, $\CI$ is
computed from the exposed classes (Eq.~\ref{eq:ci}), the question bank is built on the series the
profile keeps, the agent is run, and the four shares of Eq.~\ref{eq:decomp} are reported. The
dependence of these results on the canon is tested by recomputing the ceilings, the realized
readiness and the census under four alternative canons: without air quality, without occupancy and
lighting, with no target level above L2, and with each of the six domains counted equally. To test
whether the agent respects the ceiling, every class that a profile cannot support is also posed at
that profile, one instance with three repeats, with the question and its ground truth built on all
published series and the agent given only the profile's series.

\begin{table}[tbp]
\centering
\caption{Instrumentation profiles of the sweep. Each profile holds the stream classes of the
profiles above it and adds those listed. The last column gives which of the added classes the BATS
large office includes.}
\label{tab:profiles}
\footnotesize\linespread{1}\selectfont
\begin{tabular}{@{}l>{\raggedright\arraybackslash}p{0.35\linewidth}>{\raggedright\arraybackslash}p{0.29\linewidth}>{\raggedright\arraybackslash}p{0.15\linewidth}@{}}
\toprule
Profile & Adds these stream classes & Basis & In BATS \\
\midrule
Minimal & zone temperature & the least a zone controller reads & yes \\\addlinespace
Comfort & zone heating and cooling setpoints & a thermostat's inputs & yes \\\addlinespace
Standard & occupancy & occupancy sensing & yes \\\addlinespace
Metered & whole-building meter, end-use submeters, zone lighting power, equipment power, outside air
temperature & the building-level metering ASHRAE~90.1-2022 requires of a large building, plus
equipment power, zone lighting power and outside air temperature & all but outside air temperature \\\addlinespace
Rich & scheduled setpoint, terminal actuation command and response, supply air temperature, run status, equipment
load, zone CO$_2$, outdoor-air flow & the points the Guideline~36 sequences read and write & none \\
\bottomrule
\end{tabular}
\end{table}

\paragraph{Building and agent sweeps}\label{sec:bldg-agent}
The reference agent is also run on the other eight buildings of the corpus, on all their published
series with one repeat, to test whether the results hold across building sizes and climates. To test
how much the choice of agent matters, the same questions are run on the sweep building at the
metered profile, one repeat of the 22 classes that made up the question bank when the comparison was
run, with a second current open-weight model, Gemma~4 with 31 billion parameters~\cite{gemma4}, and
with three other local models (Qwen2.5 32B, Llama~3.1 8B and Mistral NeMo 12B), each pinned to an
exact build. The building, the questions, the tools and the grading are held fixed across models, so
a difference between models is AI failure.

\paragraph{Diagnosis scenarios}\label{sec:faults}
Diagnosis questions need faults with known causes, and a test of attribution needs pairs of faults
that look alike without the separating stream. Four scenarios in two pairs are used, all in the
large office in Tampa (\ref{app:diag}). In pair~A, a thermostat offset of 1\,\si{\celsius} in one
zone, injected through the EnergyPlus fault model~\cite{zhang2017faultmodel} so that it propagates
through the control loop and the physics, is set against a real schedule change of the same size.
The two move the zone temperature and the setpoint in effect alike, and only the scheduled setpoint
separates them (Table~\ref{tab:causes}). In pair~B, a sensor drift of 1\,\si{\celsius}, a magnitude
the sensor-fault literature treats as material~\cite{du2007sensorfault}, is set against an
internal-load step of 40\,\si{\kilo\watt} in the same zone. The drift is added to a monitoring
sensor separate from the controlling thermostat, so it never enters the control loop, and the study
makes that assumption. The two raise the reported temperature alike, and only the terminal damper
signal separates them, because it stays flat under the drift and opens under the load.

The published release exports neither separating stream. A supplement to the release, deposited with
the study's code~\cite{bara2026}, provides the scheduled setpoint and the damper position, with the
series of the two rival scenarios. \ref{app:diag} describes how the supplement was produced and
gives the checks that the two scenarios of each pair look alike without the separating stream and
differ with it.

Each scenario is asked with its candidate causes listed in the question. For pair~B the question
also states that the reported temperature comes from a monitoring sensor and that the zone's
controller uses its own thermostat, the premise under which the damper position separates the
causes. A scenario is asked in two conditions, with 20 repeats each. In the base condition the agent
has only the zone temperature and setpoint series, the causes are not identifiable, the correct
outcome is an abstention, and any answer is an overclaim (Table~\ref{tab:outcomes}). In the
separating condition the stream that separates the pair is added, the question is answerable, and
the correct outcome is the true cause, so a wrong cause or an abstention is an AI failure.

\subsection{The real buildings}
\label{sec:real}
Two public sources supply real buildings. Mortar~\cite{fierro2018mortar,mortar_hf} publishes 44
Brick KGs of real buildings with their series, and 36 of them link at least one point to a series.
The KGs are read at their revision of 11 August 2026 (commit 8844574c). The Building TimeSeries
(BTS) dataset~\cite{bts2024} adds one office, Site~B, with a Brick KG and metered series. For each
KG the stream classes are read by a fixed set of rules that map Brick classes, together with what a
point is attached to (a zone, a terminal unit, central equipment or the building), to the 17 stream
classes of the canon (Table~\ref{tab:brickmap}). A point typed with the successor of a listed class
counts as that class. A metered point that the KG attaches to nothing and links to nothing counts as
no stream class, because its scope is unknown.

\subsection{Experiments on the real buildings}
\label{sec:real-exp}
\paragraph{Census}\label{sec:census}
Both ceilings then follow from Eqs.~\ref{eq:ci} and~\ref{eq:cact}. Two answerable-readiness ceilings
are computed per KG, one from every typed point the KG declares and one from the points that carry a
series reference, and their difference is the declared-versus-linked gap. The census reads linked
points, not usable series, because usability can be judged only from each building's series, which
are downloaded for the eight buildings run in depth. The ranking robustness of
Section~\ref{sec:canon} is tested by drawing 2,000 random weightings, one independent exponential
draw per domain, and counting how many reorder at least one pair of buildings whose ceilings differ
(Eq.~\ref{eq:cw}). The census is repeated with the unattached meters counted as whole-building
meters, and with a strict mapping that credits no generic class through what it is attached to. As
an external check, each of the ten Mortar analytics applications that compute something is run on
each KG by its own qualifying queries, and the number of applications that can run is compared with
the ceiling.

\paragraph{Questions on real data}\label{sec:realq}
The question bank is run on the eight Mortar buildings with the most linked streams, each on the
calendar year in which the most of its series are usable. A series is \emph{usable} when it covers
nearly every hour of the year and is neither stuck nor dead, and only usable series are kept, so the
ceiling of a real building is computed on the data an agent could use. \ref{app:realdata} states the
rule, its sensitivity variants, and the handling of temperature units. The KGs give no time zone, so
an hour window refers to Coordinated Universal Time (UTC), and on these buildings it tests the
computation, not an occupied period. The questions and their ground truth are generated as on the
simulated corpus (Section~\ref{sec:bank}), from the building's own KG and usable series. On real
data the ground truth is what the recorded series say, so the questions test the agent's computation
and not the accuracy of the building's sensors. Each class yields up to three instances, with three
repeats on two buildings and one on the rest. \FloatBarrier

\section{Results}
\label{sec:results}
The results follow the experiments of Figure~\ref{fig:design} in order, with the actuation axis
reported after the census, and close with the reference agent's misses across all runs.

\subsection{The instrumentation sweep}
\label{sec:res-sweep}
\begin{figure}[tbp]
\centering
\includegraphics[width=0.9\linewidth]{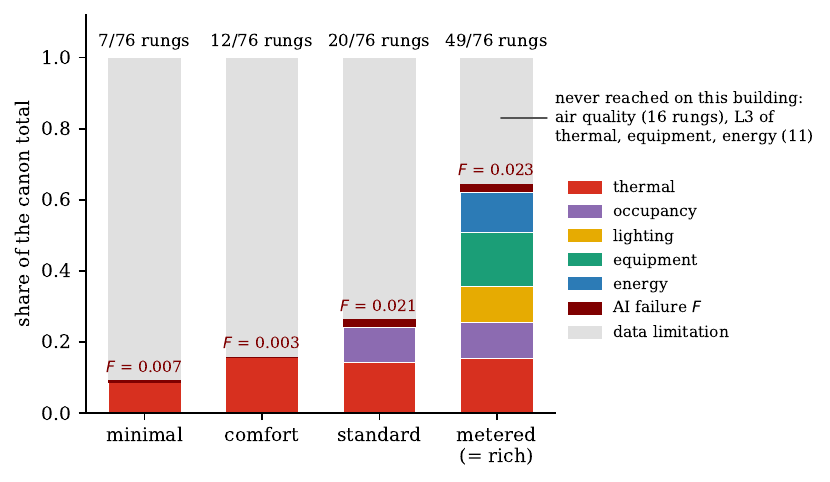}
\caption{The instrumentation sweep on the large office in Tampa, by domain, reference agent. The
coloured segments are the realized readiness of each domain.}
\label{fig:decomp}
\end{figure}

Table~\ref{tab:sweep} and Figure~\ref{fig:decomp} report the sweep. The ceiling rises from 7 to 49
of the 76 rungs as sensing is added. The metered profile reaches $\CI=0.645$, which is also the most
the published corpus can reach (Section~\ref{sec:sweep}). Air quality (16 rungs), the L3 rungs of
thermal and equipment (8 rungs), and the L3 rungs of energy (3 rungs, which need outside air
temperature) stay in the data-limitation share at every profile. Table~\ref{tab:cells} gives the
achieved level of every cell at every profile.

\begin{table}[!htbp]
\centering
\caption{The instrumentation sweep on the large office in Tampa, reference agent. $n$ is the number
of graded episodes, $\aifail/\CI$ the AI failure as a share of the ceiling, and the interval the
95\% bootstrap interval on $\Areal$. The unprobed share is zero at every profile, so AI failure is
the rest of the ceiling, $\aifail=\CI-\Areal$.}
\label{tab:sweep}
\footnotesize
\setlength{\tabcolsep}{4pt}
\begin{tabular}{@{}lrrrrrrl@{}}
\toprule
Profile & $\CI$ & $n$ & $\Acc$ & $\Areal$ & $\aifail$ & $\aifail/\CI$ & 95\% interval \\
\midrule
minimal & 7/76 & 72 & 0.921 & 0.085 & 0.007 & 0.08 & 0.080--0.089 \\
comfort & 12/76 & 114 & 0.981 & 0.155 & 0.003 & 0.02 & 0.152--0.158 \\
standard & 20/76 & 186 & 0.919 & 0.242 & 0.021 & 0.08 & 0.233--0.251 \\
metered = rich & 49/76 & 456 & 0.964 & 0.621 & 0.023 & 0.04 & 0.610--0.632 \\
\bottomrule
\end{tabular}
\end{table}

The agent realizes most of what the ceiling credits. Its accuracy over the probed rungs is 0.92 at
the minimal profile and 0.96 at the metered profile, where it realizes 0.621 of the 0.645 that the
building can support. The AI-failure share never exceeds 0.023 of the canon. Measured against the
ceiling, the agent loses 0.08 of what the data can answer at the minimal profile and 0.04 at the
metered profile. Every credited rung is probed, so realized readiness equals the product
$\CI\,\Acc$. Across the three repeats, realized readiness at the metered profile ranged from 0.599
to 0.638, and accuracy from 0.93 to 0.99. The interval of Table~\ref{tab:sweep} is narrower because
it is for the mean of the three repeats, and a single run varies by this much. Accuracy is lower at
the standard profile than at the comfort profile (0.92 against 0.98), although both pose the same
thermal questions on the same graph. Seven of the eleven thermal misses at the standard profile,
against one at the comfort profile, are abstentions after the agent failed to find a zone by its
label (Fisher's exact test on the thermal misses, $p=0.02$), so the difference lies in how the agent
searched.

At the metered profile, 40 of the 51 classes scored 1.00 in every repeat. Most of the other 11
missed one of their nine answers, and only one changed by more than one instance between repeats,
the equipment explanation at L2, which scored 1 of 3 in one repeat and 3 of 3 in another.
Table~\ref{tab:bank} gives the score of every class, among them the two-hop classes and those that
reproduce Mortar analyses.

Of the 16 misses at the metered profile, 5 are abstentions on answerable questions, 9 are wrong
answers, and two reached the turn limit without an answer.

\paragraph{Questions the data cannot answer}
Asked the classes that each profile cannot support, the reference agent answered 8 of 128 questions
at the minimal profile (overclaim rate 0.06, 95\% interval 0.03--0.12) and none at the comfort (0 of
113) or standard (0 of 90) profiles. These counts leave out one server error at each of the minimal
and comfort profiles, and five episodes ended without an answer. Every one of its 318 abstentions
named a stream the profile lacks. All eight were setpoint questions asked without a setpoint series,
and the agent answered from the static design value that the published KG stores on each zone
setpoint. The answer was right in four of the eight, where the setpoint in effect equalled the
design value, and wrong in the other four. All eight count as overclaims, because a design value
does not determine the setpoint in effect at a given hour. At the metered profile every class the
published series support is answerable, so none is posed there.

\paragraph{A guessing baseline}
A fixed answer chosen in advance, zero for a number, false for a yes/no question and the most common
name for an entity, would pass 0.33 of the 456 questions at the metered profile, against 0.965
answered correctly by the reference agent. On this building seven classes have the same answer, or
one within the tolerance, in every instance, so a constant passes all of them. Five are diagnosis
and explanation classes whose answer names the same zone, chiller or direction of change at every
sampled time. Two are thermal deviations that stay near zero because the simulated zones track their
setpoints. On these classes the score cannot tell an answer from a guess. The agent was not
guessing, however, as it abstained in 318 of the 331 questions the data cannot answer.

\paragraph{Links recorded from one side}
On the four classes run on the original KG, without the reverse links added in
Section~\ref{sec:substrate}, the reference agent answered 36 of 36, as it did with them. For this
agent the one-sided links cost nothing on those classes, so the added links do not inflate their
score.

\paragraph{Sensitivity to the canon}
Absolute values depend on the canon, and the conclusions drawn from them do not
(Table~\ref{tab:canon-sens}). Under the four alternatives, the profiles keep their order, and the
simulated large office stays far above the median real building. AI failure stays between 0.034 and
0.040 of the ceiling, and the ranking of the 37 real buildings keeps a rank correlation of at least
0.97 with the declared canon.

\begin{table}[!htbp]
\centering
\caption{Headline results under alternative canons: the ceiling and $\Areal$ of the large office at
its metered profile, and the median ceiling of the 37 real KGs with linked series with the Spearman
rank correlation of their ceilings against the declared canon.}
\label{tab:canon-sens}
\footnotesize
\setlength{\tabcolsep}{4pt}
\begin{tabular}{@{}lrrrrrr@{}}
\toprule
Canon & $\cS$ & $\CI$ & $\Areal$ & $\aifail/\CI$ & Real median & Rank correlation \\
\midrule
declared & 76 & 0.645 & 0.621 & 0.036 & 0.158 & 1.000 \\
no air quality & 60 & 0.817 & 0.787 & 0.036 & 0.200 & 0.994 \\
no occupancy, lighting & 60 & 0.550 & 0.528 & 0.040 & 0.200 & 0.999 \\
no level above L2 & 61 & 0.803 & 0.774 & 0.036 & 0.197 & 0.970 \\
equal domain shares & 76 & 0.708 & 0.684 & 0.034 & 0.125 & 1.000 \\
\bottomrule
\end{tabular}
\end{table}

\subsection{The building and agent sweeps}
\label{sec:res-others}
\paragraph{Building sweep}
On the other eight buildings of the corpus, the reference agent's accuracy ranges from 0.910 to
0.959 (Table~\ref{tab:others}). The ceiling cannot change with climate, because every climate
publishes the same streams, so the large and medium offices reach the sweep building's 49 of 76, and
accuracy stays within that narrow range. The small offices reach 37, because no equipment power is
metered in their release, and the equipment domain falls into the data-limitation share. Every
domain scores at least 0.88 in every building except occupancy in two, 0.75 on the small office in
New York (18 of 24 episodes) and 0.83 on the medium office in Buffalo (20 of 24).

\begin{table}[!htbp]
\centering
\caption{The other eight buildings of the published corpus, reference agent, 1 repeat, 113 to 153
graded episodes each. Air quality has no stream in the release, and a dash marks a domain with no
instance.}
\label{tab:others}
\footnotesize
\setlength{\tabcolsep}{3pt}
\begin{tabular}{@{}lrrccccc@{}}
\toprule
Office, climate & $\CI$ & $\Acc$ & Thermal & Energy & Equipment & Occupancy & Lighting \\
\midrule
Large, New York  & 49/76 & 0.952 & 0.95 & 0.93 & 0.97 & 0.96 & 0.96 \\
Large, Buffalo   & 49/76 & 0.932 & 0.95 & 0.93 & 0.89 & 0.96 & 0.96 \\
Medium, Tampa    & 49/76 & 0.940 & 0.95 & 0.93 & 0.88 & 0.96 & 1.00 \\
Medium, New York & 49/76 & 0.957 & 0.92 & 1.00 & 0.97 & 0.92 & 1.00 \\
Medium, Buffalo  & 49/76 & 0.924 & 0.95 & 1.00 & 0.89 & 0.83 & 1.00 \\
Small, Tampa     & 37/76 & 0.959 & 0.94 & 0.93 & --   & 0.96 & 1.00 \\
Small, New York  & 37/76 & 0.910 & 0.94 & 0.93 & --   & 0.75 & 1.00 \\
Small, Buffalo   & 37/76 & 0.955 & 0.97 & 1.00 & --   & 0.92 & 0.92 \\
\bottomrule
\end{tabular}
\end{table}

\begin{figure}[tbp]
\centering
\includegraphics[width=0.95\linewidth]{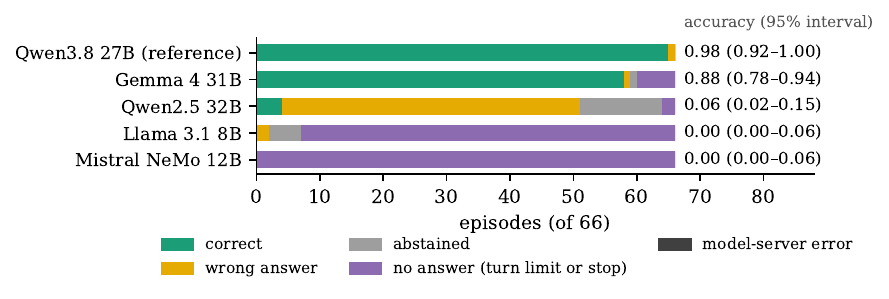}
\caption{Other agents on the sweep building, metered profile, 22 classes, 1 repeat (66 episodes
each). The interval is the Wilson 95\% interval on accuracy.}
\label{fig:agents}
\end{figure}

\paragraph{Agent sweep}
\label{sec:res-agents}
Figure~\ref{fig:agents} compares the four other local models of Section~\ref{sec:bldg-agent} with
the reference agent, whose row is the first repeat of the sweep on the same 22 classes (66
episodes). Gemma~4 reaches 0.879, against 0.985 for the reference, and takes more than three times
as long per episode (242\,s against 70\,s). Six of its eight misses ended without an answer. Qwen2.5
answers most questions and gets 4 of 66 right. Llama~3.1 and Mistral NeMo score zero because they
almost never call the answer tool. Gemma~2 9B could not be run, because the model server offers no
tool calling for it. On one building and one set of questions, accuracy thus spans 0 to 0.985 across
five local models. A readiness report must name the agent that realizes it, because the realized
share belongs to the building and the agent together.

\subsection{The diagnosis scenarios}
\label{sec:res-diagnosis}
\begin{figure}[tbp]
\centering
\includegraphics[width=0.9\linewidth]{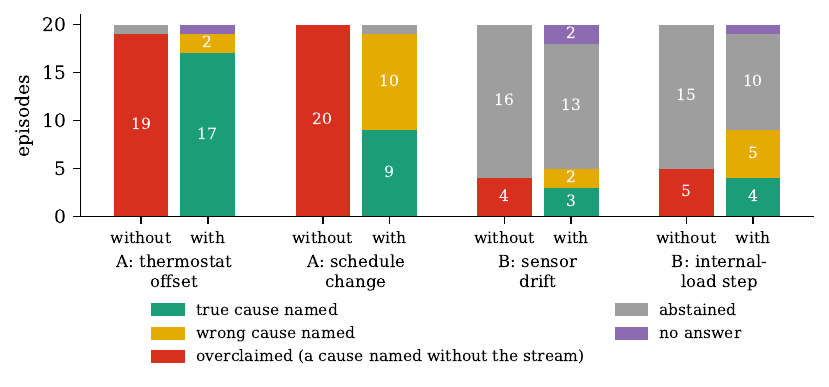}
\caption{Diagnosis outcomes, reference agent, 20 repeats per scenario and condition. Pair~B is asked
with the monitoring-sensor premise stated (Section~\ref{sec:faults}). Without the separating stream
an abstention is correct and any named cause is an overclaim. With it, only the true cause is
correct.}
\label{fig:diag}
\end{figure}

Each of the four scenarios is asked 20 times in each condition (Section~\ref{sec:faults}). Without
the separating stream the reference agent overclaimed in 48 of 80 episodes (95\% interval
0.49--0.70). With it, the agent named the true cause in 33 of 80 (0.31--0.52), a wrong cause in 19,
abstained in 24, and ended without an answer in 4 (Figure~\ref{fig:diag}). The data suffice in all
four scenarios. A fixed rule that asks only whether the separating stream changed, the scheduled
setpoint for pair~A and the damper position for pair~B, names the true cause of each
(Table~\ref{tab:pairs}). The agent did not use the stream in either pair. In pair~A it named a cause
in 39 of 40 episodes without the scheduled setpoint, most often the controller offset, and it kept
favouring the offset with the setpoint. It named the true cause of the schedule change in 9 of 20
episodes with the stream and in 9 of 20 without it (Fisher's exact test, $p=1$). In pair~B it named
the true cause in 7 of 40 episodes with the damper position and in 2 of 40 without it ($p=0.15$). In
53 of its 54 abstentions on pair~B it asked for a second temperature reading, although the question
stated that the controller's thermostat is not in the data. A first run of pair~B without that
premise had shown an effect on the load step, 11 of 20 correct with the damper position against 1 of
20 without it. With the premise stated, the effect did not hold (4 of 20 against 2 of 20).

\subsection{The census of public real-building KGs}
\label{sec:res-real}
\begin{figure}[tbp]
\centering
\includegraphics[width=0.85\linewidth]{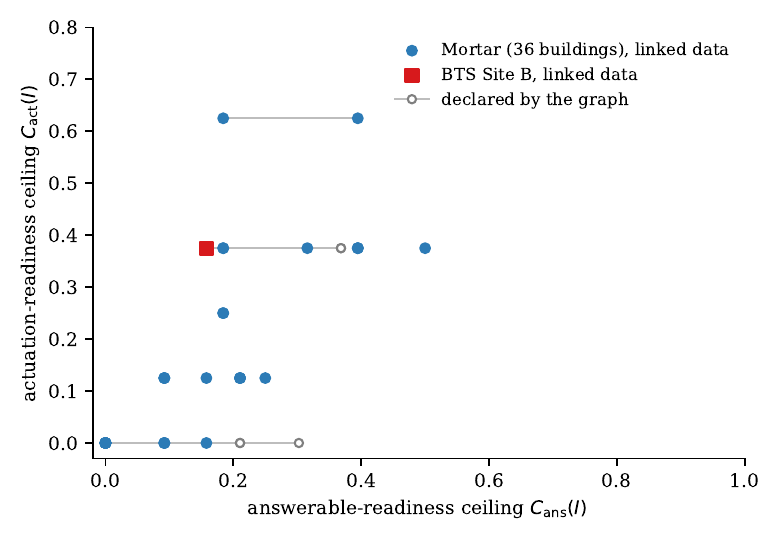}
\caption{Real buildings in the plane of the two ceilings, 37 KGs with linked series. Filled markers
use the points that carry a series. An open marker, joined by a grey line, shows the higher ceiling
the KG declares. Markers overlap where KGs share both ceilings.}
\label{fig:real-plane}
\end{figure}

\begin{figure}[tbp]
\centering
\includegraphics[width=0.9\linewidth]{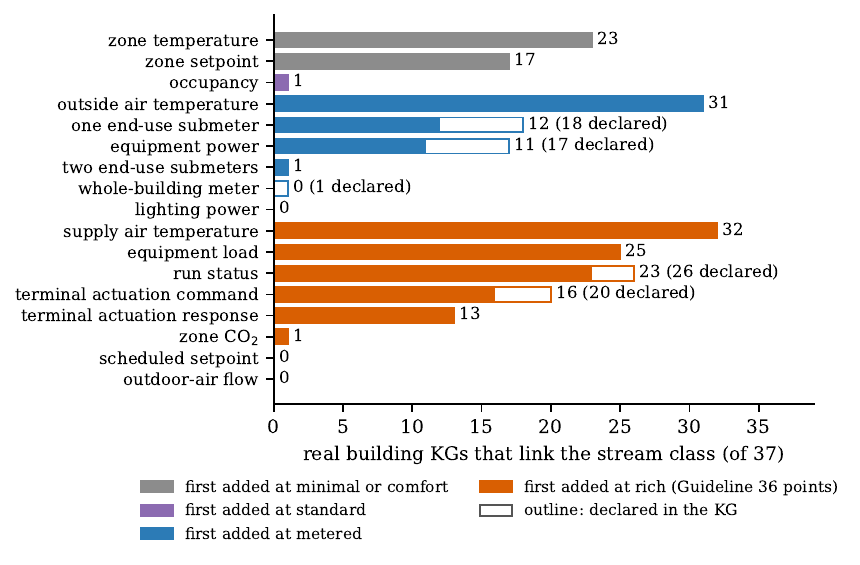}
\caption{Stream classes linked by the 37 real building KGs with linked series, coloured by the
profile of the sweep that first adds each class. An outline shows the KGs that declare the class,
where more declare it than link it.}
\label{fig:real-streams}
\end{figure}

Figure~\ref{fig:real-plane} places the 37 real building KGs that link at least one point to a series
in the plane of the two ceilings, and Table~\ref{tab:census} summarizes the census. On the data that
is linked, the answerable-readiness ceiling has a median of 0.158 and a maximum of 0.500, against
0.645 for the simulated large office at its metered profile. A few stream classes set the
distribution (Figure~\ref{fig:real-streams}). The real KGs often link the control points of the rich
profile, such as supply air temperature (32 of 37), equipment load (25) and run status (23). They
rarely link the meters of the metered profile. Twelve link an end-use submeter, 11 equipment power,
and none a point typed as a whole-building meter. One links occupancy and one CO$_2$. Some classes
are declared more often than they are linked. The terminal actuation command is declared in 20 KGs
and linked in 16, run status in 26 and 23, equipment power in 17 and 11, an end-use submeter in 18
and 12, and a whole-building meter in one KG that links none (Figure~\ref{fig:real-streams}). The
metering that the simulated office has at its metered profile is therefore rare in these KGs. The
thermal and equipment domains carry the ceiling of almost every real building, and the energy,
occupancy, air-quality and lighting domains sit in its data-limitation share.

\begin{table}[tbp]
\centering
\caption{The real-building census: 45 KGs, 37 of them with linked series. ``Declared''
counts every typed point; ``linked'' counts points that carry a series reference.}
\label{tab:census}
\small
\begin{tabular}{@{}lr@{}}
\toprule
Quantity & Value \\
\midrule
Graphs in the census (Mortar + BTS) & 45 \\
Graphs with at least one linked series & 37 \\
$\CI$ on linked data: minimum / median / maximum & 0.000 / 0.158 / 0.500 \\
Graphs whose declared $\CI$ exceeds the linked $\CI$ & 15 of 45 \\
$\Cact$: minimum / median / maximum & 0.000 / 0.125 / 0.625 \\
Graphs with no actuation handle & 13 of 37 \\
Random domain weightings that reorder at least one pair & 1,901 of 2,000 \\
Pairs of buildings that ever change order & 102 of 575 \\
\bottomrule
\end{tabular}
\end{table}

\paragraph{Declared against linked}
In 15 of the 45 KGs the ceiling computed on every typed point exceeds the ceiling computed on the
points that carry a series. In nine of them the linked ceiling is zero. Seven of these link no point
to a series at all, and two link only series that reach no rung of the canon. The eighth KG without
any linked series declares no canon stream either, so both its ceilings are zero. The simulated
corpus shows the same gap, since the published large-office KG declares 74 of the 76 rungs and its
series support 49.

\paragraph{Ranking under other weights}
Of 2,000 random weightings of the six domains, 1,901 reorder at least one pair of the 37 buildings.
Most pairs are stable, and only 102 of the 575 pairs with unequal ceilings change order under any
weighting. A site that weights the domains must therefore report the weights with the ranking.

\paragraph{Ceiling against a sensor count}
A ceiling that only counted sensors would add nothing to a point list. Over the 37 KGs, the linked
ceiling has a rank correlation of 0.61 with the number of linked points, and 145 of the 573 pairs
that both measures separate are ordered the opposite way. Its rank correlation with the number of
linked stream classes is 0.83, and 64 of the 539 pairs that both measures separate are reversed.
Graphs that link six stream classes range from 0.158 to 0.211. The ceiling therefore follows which
classes are linked and how they combine in the requirement table, not how many points or classes
there are.

\paragraph{Sensitivity to the mapping rules}
The census depends on the rules of Table~\ref{tab:brickmap}, which read a KG's structure and never
its point names. Counting the 16 retyped building meters of the Mortar graphs as whole-building
meters leaves the median at 0.158, with a rank correlation of 0.98 against the reported ceilings. A
strict mapping that credits no generic class through what it is attached to lowers the median to
0.092, with a rank correlation of 0.65, so the attachment rules matter more than any single class
(\ref{app:realdata}).

\paragraph{Ceiling against applications that can run}
Only three of the ten Mortar applications can run on any KG, and the number that can run has a rank
correlation of 0.62 with the linked ceiling, and 0.75 on the June revision of the graphs. The
applications cover thermal comfort and setpoint tracking, so this check covers mainly the thermal
part of the ceiling (\ref{app:realdata}).

\subsection{Actuation-readiness}
\label{sec:results-act}
The actuation-readiness ceiling ranges from 0 to 0.625, with a median of 0.125, and 13 of the 37 KGs
expose no handle of the actuation canon at all. Among the 37 KGs, equipment is the domain most often
actuable (20 KGs, almost all through an on-off or enable command at A1), thermal next (17 KGs, 16 of
them at A2 through a terminal command), air quality through ventilation rarely (3 KGs), and lighting
never. No KG reaches the full ceiling of 1.0, and the actuation axis is only loosely related to the
answering axis. The five buildings with the same linked answerable ceiling of 0.184 range from 0.25
to 0.625 on the actuation axis. On the simulated large office the actuation ceiling is 0.75, with
thermal, air quality and equipment at A2 and no lighting handle, against an answerable ceiling of
0.645. That value counts the command points its KG declares, because the release exports none of
their series, so it is not comparable with the linked values of the real buildings. The caveat of
Section~\ref{sec:actuation} applies with more force to the real KGs. A command point in a Mortar KG
shows that the point is modelled, not that an external agent may write to it.

\subsection{Questions on real data}
\label{sec:res-realq}
On the eight Mortar buildings with the most linked streams, 2,683 linked series of the point classes
that the questions need were converted, and 2,008 of them (75\%) passed the usability rule of
Section~\ref{sec:realq}. None was absent from the release, so every loss is a series that is present
but unusable, and the losses fall on particular classes. Every heating command (0 of 181) and every
occupancy sensor (0 of 104) failed, as did 213 of 803 zone temperature sensors, while zone setpoints
(622 of 700), damper setpoints (215 of 241), supply airflow (275 of 307) and outside air temperature
(42 of 44) mostly passed. The two classes fail for reasons no threshold repairs. The heating
commands are sampled about once every eight hours, and holding each value until the next still
leaves none usable. Every occupancy series holds a single value over the year. Across the variants
of the rule, between 58\% and 85\% of the linked series are usable, and none makes a heating command
or an occupancy sensor usable. A census on declared points would credit one of these buildings with
the occupancy domain, and its data cannot support it.

On the eight buildings, the question bank yields questions for the thermal classes only, because
none of the eight links a usable meter, airflow setpoint or heating command. Thirteen classes
instantiate, among them the four that judge a zone against its setpoint on the seven buildings that
link a usable one and the L3 comparison of terminal damper commands on the buildings that link both
the command and the response. They yield 394 episodes. The reference agent scored 375 of the 394
(0.952, 95\% interval 0.93--0.97). Building accuracy ranges from 0.818 to 1.000, and the two
buildings run three times range from 0.939 to 0.970 and from 0.917 to 0.972 across repeats. The
weakest class is the setpoint diagnosis, the zone farthest above its own setpoint, at 20 of 30,
mostly episodes that reached the turn limit. Of the 19 misses, nine ended without an answer, four
name the room by its full point path, four are wrong or misplaced answers, and two are abstentions.
On the thermal questions that the real data support, the agent is therefore about as accurate as in
simulation (0.952 against 0.96 on the same thermal classes at the metered profile). What real data
lowers is the number of questions that can be asked.

\subsection{The reference agent's misses}
\label{sec:res-misses}
Over the 2,380 graded episodes of the reference agent on the published corpus and the real
buildings, 120 were misses. Each was classed by fixed rules and the rules were checked by hand
(Table~\ref{tab:misses}). All but four are the agent's own errors, and the largest group is one
pattern: a question names a zone by its label, the agent searches the identifiers, finds no match,
and abstains although the data are present. A second group gives the right value in the wrong field
of the answer. The other four are real-building answers that name a room by its full point path,
which a looser grader would accept. All 120 are scored as AI failure in the results above, so the
AI-failure share is an upper bound on the errors that belong to the agent.

\begin{table}[!htbp]
\centering
\caption{The 120 misses of the reference agent over 2,380 graded episodes (the sweep, the
other eight buildings, the runs on the original KG and the real buildings).}
\label{tab:misses}
\footnotesize
\begin{tabular}{@{}lrp{0.58\linewidth}@{}}
\toprule
Class & Count & What happened \\
\midrule
agent error & 116 & 44 abstentions although the data were present, most after failing to find a
zone named by its label; 20 episodes without an answer at the turn limit; 14 right values placed
outside the answer field; 13 occupied and unoccupied swapped; 25 other wrong values, entities,
signs or units \\
grader strictness & 4 & a real-building room named by its full point path \\
\bottomrule
\end{tabular}
\end{table}

\FloatBarrier

\section{Discussion}
\label{sec:discussion}

\subsection{Instrumentation bounds reporting and judging}
\label{sec:disc-lever}
On the questions a building's data can answer at L1 and L2, the reference agent answers nearly all
of them, and the AI-failure share stays at or below 0.023 of the canon at every profile of the
sweep. This accuracy is far above the top row-matching $F_1$ of 0.38 that zero-shot systems reach on
the BuildingQA benchmark~\cite{mulayim2025buildingqa}. The two numbers measure different tasks and
are not directly comparable, but much of the gap lies in how the questions are posed. BuildingQA
poses paraphrased natural-language questions whose answers are the rows of a KG query. Here every
question states its conventions and answer shape, the agent can query and compute as often as it
needs within 25 turns, and an entity may be named by any name the KG gives it. The accuracy reported
here is therefore accuracy on well-posed questions, and that design is what isolates the data as the
constraint. What moves realized readiness is the ceiling, which rises from 0.09 to 0.64 as sensing
is added, and what caps it on the published corpus is the absence of the control points that
Guideline~36 names. For practice, ``is this building ready for AI?'' is to first order ``what can
this building's data answer?'', which can be computed before any AI is run. Naming a cause is the
exception (Section~\ref{sec:disc-diag}). Because the ceiling rises in steps, a readiness audit
yields a ranked list: the stream class that would raise a given building's ceiling the most. For the
simulated office at its metered profile the largest single step is air quality, where a CO$_2$
sensor per zone~\cite{pei2019dcv}, with occupancy and outdoor-air flow, opens 16 rungs. For 26 of
the 37 real buildings with linked data it is equipment power, which 11 link, for ten it is a
whole-building meter, which none links as such, and for one it is CO$_2$.

The list also has a cost axis. A stream class can be added by installing a sensor or by linking a
point that the automation system already records, and the census shows that the second kind is
common: 15 of 45 KGs declare more than their linked data support, and on the eight buildings studied
in depth 25\% of the linked series are present but unusable. Linking and cleaning cost labour and no
hardware. The retrofit list should therefore be read in two passes, first the points to link and
repair, then the sensors to install, and the ceiling can be recomputed after each pass from the KG
alone. The same decomposition gives the return on each kind of application. Reporting and summary
applications need the L1 and L2 rungs, which zone temperature, setpoints and a meter supply.
Diagnostic applications need the L3 rungs and so the control points. Control applications need the
actuation axis. A building can be ready for reporting and not for diagnosis. In the census, 17 of
the 37 buildings with linked data link a zone setpoint and none links a point typed as a
whole-building meter, so most are not yet ready even for reporting across all domains.

\subsection{When the KG is incomplete or wrong}
\label{sec:kg-quality}
The ceiling is read from the KG, so it inherits the KG's errors, and different errors bias it in
different directions. Points that the KG declares but that carry no data make it overstate what the
building can answer (Section~\ref{sec:res-real}). Points that the automation system records but that
the KG never models make it understate, and they form the first pass of the retrofit list
(Section~\ref{sec:disc-lever}). Mistyped or renamed classes make the KG misreport what the building
has, and the error is silent. In the Mortar graphs, 1,084 zone setpoints carry renamed Brick classes
that rules knowing only the old names would miss without warning, and the census counts them by
following Brick's links to each successor. The 16 building meters are retyped as generic sensors
that no rule reading the KG's structure can recognize. Wrong or one-directional relations leave the
data in place but make them harder to reach, so the loss appears as AI failure, not as data
limitation. On the four classes tested, one-directional links did not affect the reference agent,
but a less thorough agent or a sparser KG could be affected. A KG can also mislead an agent into
answering, as the design values stored on the published KG's setpoints did at the minimal profile
(Section~\ref{sec:res-sweep}). The same errors move the actuation ceiling. A handle that the KG does
not model lowers $\Cact$, and a handle that it models but the automation system does not let an
agent write to raises it (Section~\ref{sec:actuation}). The measurement is therefore only as good as
the KG, and an assessment should begin by comparing the declared, linked and usable ceilings before
relying on any one of them.

\subsection{Diagnosis and the limits of published data}
\label{sec:disc-diag}
Answerability at L3 is unique determinability (Section~\ref{sec:identifiability}). A richly
instrumented building can still be unable to answer ``why'', and sensor counts and checklist scores
cannot see this, because they credit the presence of a related stream. The requirement table follows
from the cause table of \ref{app:canon}, so a change in the declared causes changes the requirement
in a way a reader can audit.

Without the separating stream the agent overclaimed in most diagnosis episodes
(Section~\ref{sec:res-diagnosis}). An agent that names a cause the data cannot support is a hazard
in a diagnostic role whatever its accuracy elsewhere, so the overclaim rate belongs beside the
accuracy in any readiness report. With the stream the data suffice, and the agent still did not use
it. The framework scores those outcomes as AI failures, which is the distinction it exists to draw.
At L1 and L2 the data bound readiness. At L3 the data are necessary but not sufficient, and for this
agent the AI is the constraint.

\subsection{Relation to the Smart Readiness Indicator}
The framework offers the kind of measurement that the EU Smart Readiness Indicator's own methodology
calls preferable and does not provide: a rating grounded in what the building's data can answer,
computed from the KG and the series~\cite{verbeke2020sri}. The two instruments are complementary and
meet in one place. Ten of the SRI's 54 services report information, and five of them, on heating,
cooling, indoor air quality, electricity consumption and central performance reporting, relate to
the canon through data. Their upper functionality levels, such as benchmarking or fault detection,
are awarded by an assessor on inspection, and the ceiling tells whether the streams behind them
exist and hold usable data. The crosswalk of \ref{app:sri} maps every service to a canon cell, to
the actuation axis, or to neither. The SRI weights its domains by an energy balance, and most
weightings reorder at least one pair of real buildings (Section~\ref{sec:res-real}), so a site that
adopts the SRI's weights should report the ceiling under those weights and with every rung counted
equally (Eq.~\ref{eq:ci}) together.

\subsection{Two axes, and the gradable boundary of autonomy}
Readiness for autonomy has two prerequisites a building either affords or does not: the data to
answer and the handles to act. Both are measured as coverage, $\CI$ and $\Cact$, and the measurement
stops short of the quality of what an agent then does, because a control policy has no objective
ground truth. Simulation benchmarks such as the Building Optimization Testing framework
(BOPTEST)~\cite{blum2021boptest} do score control policies, against performance indicators in a
model of the building, but that scoring needs a calibrated model of the building under test, which a
readiness measurement read from the KG does not have.

That the two axes are decoupled in the real buildings (Section~\ref{sec:results-act}) is itself an
operational finding, because each mismatch of Section~\ref{sec:actuation} implies a different
retrofit. The observable-but-not-actuable building needs actuators, and the
actuable-but-not-observable building needs sensors before its actions can be trusted. Two extensions
could raise readiness without new hardware: virtual sensing~\cite{li2022virtualsensor}, which infers
an unmetered quantity from correlated streams, on the answering axis, and surfacing already-present
but unmapped command points on the actuation axis.

\subsection{An evaluation pathway for building digital twins}
A building digital twin couples a physical asset to a continuously updated virtual
model~\cite{su2025dtreview}, increasingly a semantic one in which sensor streams are linked through
a KG~\cite{boje2020dt} and integrated with the automation system for live
control~\cite{ma2025dtbuilding}. Such a twin is the natural substrate for the AI-driven operation
this study concerns, and its fitness for that role is bounded by the same property measured here.
The framework therefore doubles as an evaluation pathway for a twin's answerability layer, with
$\CI$ scoring how much of the question canon the twin can support and the decomposition placing any
shortfall as missing instrumentation, unprobed credit, or agent error. A deployed twin adds live,
bidirectional synchronization with its physical counterpart, which this study does not exercise.

\subsection{Limitations}
\label{sec:limitations}
The evidence has clear edges. \emph{(i)~The controlled study is simulated.} The sweep, the diagnosis
scenarios and the model comparison run on EnergyPlus output. The real buildings supply ceilings,
usability and a subset of the question bank, and they show that the simulated profiles are richer
than most of the public real-building KGs, not that the simulated accuracy transfers. \emph{(ii)~One
reference agent and well-posed questions.} One pinned open-weight model is held fixed, so its
accuracy is a statement about that model on this question bank. The other models, whose accuracy on
the same questions ranges from 0 to 0.88, are a sensitivity analysis, not a survey. The questions
are also easier than those operators pose. Each states its conventions and answer shape, and the
agent computes over clean or filtered series. Open-ended, ambiguous or multi-step questions, on
which KGQA systems still fail often~\cite{mulayim2025buildingqa}, would lower the realized share and
raise the AI-failure share, and testing them is left to future work. Seven classes have the same
answer in every instance on the sweep building, so a constant would pass them, and sampling
instances whose answers differ would make them discriminating. \emph{(iii)~Nine buildings of one
family.} The published corpus holds three office prototypes in three climates at one code vintage,
so the ceilings it spans are those of code-compliant offices with metering and no control points.
\emph{(iv)~Cause attribution rests on four scenarios.} The positive test uses two pairs of scenarios
with 20 repeats per condition, in one building, and each question names only the two causes of its
pair. It shows that this agent did not use a separating stream in these scenarios, not a rate that
carries over to other faults. A first run of pair~B without the monitoring-sensor premise differed
on the load step (Section~\ref{sec:res-diagnosis}), so the result depends on how the question states
the sensor's role. The canon separates a stuck damper by the terminal command and its response, and
no scenario tests that pair. The L3 question classes test the use of the separating stream, not the
attribution. In pair~A, the EnergyPlus thermostat offset shifts the setpoint in effect, which a
biased thermostat in a real building would not show. \emph{(v)~Real data reaches few cells.}
Real-data accuracy is known only for the 13 thermal classes that instantiate on the eight buildings
run in depth (Section~\ref{sec:res-realq}). \emph{(vi)~Not every miss is the agent's.} Four
real-building misses would pass a looser grader, so the AI-failure share is an upper bound on the
agent's own errors (Section~\ref{sec:res-misses}). \emph{(vii)~Stated rules.} Several results rest
on rules the study declares: the Brick class-mapping rules that read stream classes from a KG, the
rule that zone-attached power in the published corpus is lighting, the 90\% coverage and
three-distinct-values rule for a usable series, and the Fahrenheit inference for unlabelled
temperatures. Each is in the appendix or the code, and per-building ceilings move when the rules
move (Section~\ref{sec:res-real}). \emph{(viii)~Actuation is coverage.} $\Cact$ records that a
handle is modelled, not that an external agent may write to it or that it would act well.
\emph{(ix)~Data sources beyond the KG.} Weather, tariffs, grid signals and unstructured records such
as work orders are outside the canon by design, and a building with none of them can still be fully
ready on the canon. \emph{(x)~Air quality is CO$_2$ only.} A building that logs particulate matter,
volatile organic compounds or humidity but not CO$_2$ scores zero on air quality, so $\CI$
understates it. Each added pollutant needs a stream class and a cause set of its own, which the
canon can take without changing the method. \emph{(xi)~Presence, not coverage.} A stream class
counts when one point of it carries usable data, so one CO$_2$ sensor in one zone credits every
air-quality rung. The ceiling states what can be answered somewhere in the building, not everywhere,
and the share of zones or units each class covers could be reported beside it. The series are also
hourly, which hides fault signatures shorter than an hour. \emph{(xii)~Air-side requirements.} The
thermal L3 requirements include the supply air temperature of the air system serving the zone, which
suits air-based systems. A radiant or hydronic zone reaches thermal L3 only if equivalent streams,
such as a supply water temperature, are added to the stream classes. \emph{(xiii)~The census reads
public research KGs.} The Mortar KGs were assembled for a research testbed, so the census measures
what those KGs describe and link. Its values move with the mapping rules and the dataset revision
(Section~\ref{sec:res-real}), and they describe public Brick models, not the building stock.
\emph{(xiv)~No expert elicitation.} The canon and the cause table were written for this study and
checked against two practitioner corpora, not rated by practitioners. A structured review by
practitioners is the most direct way to strengthen them.

\section{Conclusion}
\label{sec:conclusion}

This paper has argued that the first question for building AI is not how capable the model is but
how much the building's data can answer, and has made that question measurable. The framework
defines a building's answerable-readiness ceiling as a function of its instrumentation, the realized
readiness that a fixed reference agent delivers on the rungs the question bank probes, and a
decomposition of the whole canon into realized readiness, AI failure, unprobed credit and data
limitation. It adds a second axis, actuation-readiness, and derives the canon's diagnostic
requirements from declared cause sets.

The empirical picture is measured, and it is modest. On the published simulated corpus the ceiling
of a code-compliant large office reaches 0.64 of the canon at its metered profile, and the published
release supports no more, because it exports none of the control points that diagnosis needs. With a
question at every credited rung, the reference agent answers 0.96 of them at that profile, so the
measured realized readiness is 0.62 and AI failure is 0.02 of the canon. On these fully specified
questions, missing data, not the agent, set the limit. Naming a cause is where the agent, not the
data, falls short. Without the stream that separates two look-alike causes the agent names a cause
in 48 of 80 diagnosis episodes. With it, the agent names the true cause in 33 of 80, and in no
scenario does the stream change its answer significantly, although a fixed rule on that stream names
every cause. Across 37 public real-building KGs with linked data, the median ceiling is 0.16 and the
maximum 0.50, none links a point typed as a whole-building meter, and 15 of 45 KGs declare more than
their data support. On the eight real buildings run in depth, 25\% of the linked series are
unusable, and the agent's accuracy on the questions that remain, all of them thermal, is as high as
in simulation. Most reweightings of the domains reorder at least one pair of real buildings. The
actuation axis is only loosely related to the answering axis in real buildings. For practice, this
turns ``is my building AI-ready?'' into an auditable, ranked retrofit question whose first pass
costs no hardware and can be run on a building's own KG with the released code. For policy, it
offers one way to supply the measurement-based component that the Smart Readiness Indicator's own
methodology calls for.

Several directions follow. The most immediate is to test attribution across a larger fault library,
which needs a corpus that exports the control points of every building and belongs to the successor
of the corpus used here. Real buildings with linked meters and setpoints, which the census shows
exist, would extend real-data accuracy beyond the thermal domain. And because realized readiness
depends on the agent as well as the data, tracking how deployed agents close the AI-failure share
and hold the overclaim rate down on real data is its own line of study.

\availability{The simulated corpus, BATS v1.0~\cite{bats2026}, is published on Zenodo
(\url{https://doi.org/10.5281/zenodo.20777376}) under CC BY-NC 4.0. The Mortar KGs and series are
public Hugging Face datasets~\cite{mortar_hf}, and the BTS data are published on figshare
(\url{https://doi.org/10.6084/m9.figshare.28705559}). The code that computes both ceilings, builds
and grades the question bank and runs the agent is deposited on Zenodo~\cite{bara2026}
(\url{https://doi.org/10.5281/zenodo.23197610}) under the Apache-2.0 license. The same deposit
holds, under CC BY-NC 4.0, the result tables, the run summaries, the transcripts of the runs on the
simulated corpus, and a supplement to BATS with the separating streams and the paired diagnosis
scenarios. The Mortar data are not redistributed, so the runs on the real buildings are deposited as
scores only.}
\competing{The author declares no known competing financial interests or personal relationships that could
have appeared to influence the work reported in this paper.}
\funding{This research did not receive any specific grant from funding agencies in the public, commercial,
or not-for-profit sectors.}
\credit{\textbf{Wooyoung Jung:} Conceptualization, Methodology, Software, Validation, Formal analysis,
Investigation, Data curation, Writing -- original draft, Writing -- review \& editing,
Visualization.}
\aiuse{During the preparation of this work the author used Claude Opus~4.8, Claude Opus~5 and Claude
Opus~5.5 (Anthropic) in order to proofread the manuscript. After using these tools, the author
reviewed and edited the content as needed and takes full responsibility for the content of the
published article.}

\appendix
\section{The SRI services and the canon}
\label{app:sri}
Table~\ref{tab:sri} lists the services of the SRI's Method B catalogue~\cite{verbeke2020sri} that
relate to the canon. A service relates to the canon through the data it reports on or acts from (D),
or through the control handle it needs on the actuation axis (A). Of the 54 services, 16 relate
through data, 22 through actuation (8 through both), and 24 lie outside the canon by design
(domestic hot water, the envelope, generation, storage, vehicle charging, and grid interaction).

{\footnotesize\setlength{\tabcolsep}{3pt}
\begin{longtable}{@{}>{\raggedright\arraybackslash}p{0.14\linewidth}>{\raggedright\arraybackslash}p{0.31\linewidth}>{\raggedright\arraybackslash}p{0.09\linewidth}>{\raggedright\arraybackslash}p{0.40\linewidth}@{}}
\caption{The 30 SRI Method B services that relate to the canon. D = related through data; A = related through the actuation axis; TABS = thermally activated building systems; IAQ = indoor air quality; TBS = technical building systems.}\label{tab:sri}\\
\toprule
SRI domain & Service & \hspace{0pt}Relation & Canon cell or axis \\
\midrule
\endfirsthead
\toprule
SRI domain & Service & \hspace{0pt}Relation & Canon cell or axis \\
\midrule
\endhead
\bottomrule
\endlastfoot
Heating & Heat emission control & A & thermal A1 (zone setpoint), A2 (valve or reheat command) \\
Heating & Emission control for TABS & A & thermal A1 \\
Heating & Control of distribution fluid temperature & A & equipment A1 \\
Heating & Control of distribution pumps & A & equipment A2 (speed command) \\
Heating & Heat generator control (all except heat pumps) & A & equipment A1, A2 \\
Heating & Heat generator control (heat pumps) & A & equipment A2 \\
Heating & Sequencing of heat generators & A, D & equipment A2; equipment comparison L2 \\
Heating & Report information regarding heating system performance & D & equipment status, summary L1--L2; level 3--4 need equipment comparison L3 and diagnosis L3 \\
Cooling & Cooling emission control & A & thermal A1, A2 \\
Cooling & Emission control for TABS (cooling) & A & thermal A1 \\
Cooling & Control of chilled water temperature & A & equipment A1 \\
Cooling & Control of distribution pumps & A & equipment A2 \\
Cooling & Interlock: avoiding simultaneous heating and cooling & D & equipment verification L3 (\texttt{q\_verification\_3\_simul}) \\
Cooling & Generator control for cooling & A & equipment A2 \\
Cooling & Sequencing of cooling generators & A, D & equipment A2; equipment comparison L2 \\
Cooling & Report information regarding cooling system performance & D & as for heating \\
Ventilation & Supply air flow control at room level & A, D & air quality A1 (airflow setpoint), A2 (outdoor-air damper); air quality status L1--L2 (CO$_2$, occupancy) \\
Ventilation & Air flow or pressure control at the air handler & A, D & air quality A1; thermal verification L3 (\texttt{t\_verification\_3\_flow}) \\
Ventilation & Heat recovery: prevention of overheating & A & equipment A2 \\
Ventilation & Supply air temperature control at the air handler & A, D & equipment A1; SAT stream (thermal diagnosis L3) \\
Ventilation & Free cooling with mechanical ventilation & A, D & air quality A2; OAT stream \\
Ventilation & Reporting information regarding IAQ & D & air quality status, summary L1--L2; level 3 (warnings) needs air quality verification L3 \\
Lighting & Occupancy control for indoor lighting & A, D & lighting A2; lighting status L2 (\texttt{l\_status\_2}), lighting diagnosis L2 \\
Lighting & Control of artificial lighting power based on daylight & A & lighting A1 (level setpoint) \\
Electricity & Reporting information regarding electricity consumption & D & energy status, summary L1--L2; level 3--4 (appliance level, recommendations) need energy comparison L3 (two end-use groups) \\
Monitoring and control & Run time management of HVAC systems & A, D & equipment A1 (on/off, enable); equipment verification L3 (power while commanded off) \\
Monitoring and control & Detecting faults of technical building systems and support to diagnosis & D & diagnosis L3 in every domain; level 3 ("diagnosing functions") is the canon's L3 \\
Monitoring and control & Occupancy detection: connected services & D & occupancy status L1; lighting status L2 \\
Monitoring and control & Central reporting of TBS performance and energy use & D & energy summary L2 (per carrier, per domain: two end-use groups) \\
Monitoring and control & Single platform for automated control and coordination & A & the actuation axis as a whole (every domain at A2) \\
\end{longtable}}

\FloatBarrier
\section{Canon examples, cause sets, stream classes and achieved levels}
\label{app:canon}

Table~\ref{tab:canon-ex} gives an example question at each level of every canon cell, to
show what each level adds.

{\footnotesize\linespread{1}\selectfont\setlength{\tabcolsep}{3pt}
\begin{longtable}{@{}>{\raggedright\arraybackslash}p{0.13\linewidth}>{\raggedright\arraybackslash}p{0.26\linewidth}>{\raggedright\arraybackslash}p{0.26\linewidth}>{\raggedright\arraybackslash}p{0.30\linewidth}@{}}
\caption{An example question at each level of every canon cell. Each example uses only the stream
classes that Table~\ref{tab:canon-req} requires at its level. ``--'' marks a level above the cell's
target level. The questions posed to the agent are the question classes of \ref{app:bank}.}
\label{tab:canon-ex}\\
\toprule
Intent & L1 & L2 & L3 \\
\midrule
\endfirsthead
\toprule
Intent & L1 & L2 & L3 \\
\midrule
\endhead
\bottomrule
\endlastfoot
\multicolumn{4}{@{}l}{\textit{Thermal}} \\
status & What is Zone~3's temperature now? & How far is Zone~3 above its cooling setpoint now? & -- \\
summary & What was Zone~3's mean temperature yesterday? & For how many hours yesterday was Zone~3 outside its setpoints? & -- \\
comparison & Was Zone~3 warmer than Zone~5 at 15:00? & How much warmer was Zone~3 than the other zones on its air handler? & Was Zone~3 warmer than Zone~5 because its damper delivered less air? \\
verification & Did Zone~3 stay at or below \SI{24}{\celsius} all day? & Did Zone~3 stay within its setpoints all day? & When Zone~3 ran above its cooling setpoint, did its damper open to correct it? \\
diagnosis & Which zone was warmest at 15:00? & Which zone was farthest above its own cooling setpoint at 15:00? & Zone~3 is too warm: is the cause a setpoint override, a stuck damper or warm supply air? \\
explanation & How much did Zone~3's mean temperature change from Monday to Tuesday? & Did Zone~3 drift further from its setpoint on Tuesday than on Monday? & Why did Zone~3 run warmer on Tuesday: a schedule change, an override, a load change or a supply problem? \\
\addlinespace
\multicolumn{4}{@{}l}{\textit{Energy}} \\
status & How much electricity did the building use yesterday? & -- & -- \\
summary & What was the building's peak hourly demand last month? & What share of yesterday's electricity went to lighting? & -- \\
comparison & Did the building use more electricity on Tuesday than on Monday? & Was Tuesday's use unusual for a weekday this month? & Was Tuesday's increase due to hotter weather, and which end use drove it? \\
diagnosis & On which day last month was use highest? & Which days last month were more than 10\% above the month's mean? & Which end use caused the high-use days, and were they weather-driven? \\
explanation & By how much did daily use change from Monday to Tuesday? & Did the change fall in the occupied hours or outside them? & Why did use rise on Tuesday: the weather, or a change in one end use? \\
\addlinespace
\multicolumn{4}{@{}l}{\textit{Air quality}} \\
status & What is the CO$_2$ level in Zone~3 now? & Is Zone~3's CO$_2$ above 1,000~ppm while it is occupied? & -- \\
summary & What was Zone~3's mean CO$_2$ yesterday? & For how many occupied hours yesterday was CO$_2$ above 800~ppm? & -- \\
comparison & Was CO$_2$ higher in Zone~3 than in Zone~5 at 15:00? & How much higher was Zone~3's mean CO$_2$ than Zone~5's yesterday? & Was CO$_2$ higher in Zone~3 than in Zone~5 because it held more people or got less outdoor air? \\
verification & Did Zone~3 stay below 1,000~ppm all day? & Did Zone~3 stay below 1,000~ppm in every occupied hour? & Did outdoor-air flow rise when Zone~3 was occupied, as the ventilation design intends? \\
diagnosis & Which zone had the highest CO$_2$ at 15:00? & Which zone had the most occupied hours above 1,000~ppm? & Is Zone~3's high CO$_2$ due to more people than designed or to too little outdoor air? \\
explanation & How much did Zone~3's peak CO$_2$ change from Monday to Tuesday? & Did the rise fall in the occupied hours? & Why was CO$_2$ higher on Tuesday: more people, or less outdoor air? \\
\addlinespace
\multicolumn{4}{@{}l}{\textit{Equipment}} \\
status & What is AHU-1's supply fan power now? & What fraction of its own peak is AHU-1 drawing now? & -- \\
summary & What was AHU-1's mean power yesterday? & For how many hours yesterday did AHU-1 run above half its peak? & -- \\
comparison & Did AHU-1 draw more power than AHU-2 at 15:00? & Which of the three air handlers used the most electricity yesterday? & Did AHU-1 draw more than AHU-2 because it moved more air? \\
verification & Did AHU-1 exceed 20~kW at any hour yesterday? & Did AHU-1 stay below 90\% of its own peak all day? & Did AHU-1 draw power only while it was commanded on? \\
diagnosis & Which item drew the most power at 15:00? & Which item ran closest to its own peak yesterday? & Is AHU-1's high power due to a command change, a higher load or degradation? \\
explanation & How much did AHU-1's daily energy change from Monday to Tuesday? & Did the change fall in the occupied hours? & Why did AHU-1 use more on Tuesday: longer run hours, more airflow or degradation? \\
\addlinespace
\multicolumn{4}{@{}l}{\textit{Occupancy}} \\
status & How many people are in Zone~3 now? & Is Zone~3 busier now than its usual daytime mean? & -- \\
summary & What was Zone~3's peak occupancy yesterday? & Was Zone~3's working-hour occupancy yesterday above its monthly mean? & -- \\
comparison & Were more people in Zone~3 than in Zone~5 at 15:00? & On average, how many more people were in Zone~3 than in Zone~5 during working hours? & -- \\
diagnosis & At what hour was Zone~3 busiest yesterday? & For how many hours was Zone~3 occupied outside working hours? & -- \\
\addlinespace
\multicolumn{4}{@{}l}{\textit{Lighting}} \\
status & What is Zone~3's lighting power now? & Are Zone~3's lights on while it is unoccupied? & -- \\
summary & What was Zone~3's mean lighting power yesterday? & Was Zone~3's lighting energy yesterday above its usual weekday? & -- \\
comparison & Was Zone~3's lighting power higher than Zone~5's at 15:00? & Which zone used the most lighting energy yesterday? & -- \\
diagnosis & Which zone had the highest lighting power at 15:00? & For how many hours were Zone~3's lights on while it was unoccupied? & -- \\
\end{longtable}}

Table~\ref{tab:causes} lists the declared cause sets from which the L3 rows of the requirement
table (Table~\ref{tab:canon-req}) are derived (Section~\ref{sec:identifiability}).

\begin{table}[!htbp]
\centering
\caption{Declared cause sets and the stream that separates each cause from its rivals. ``T
(peers)'' is the zone temperature of other zones on the same air system.}
\label{tab:causes}
\footnotesize\linespread{1}\selectfont
\setlength{\tabcolsep}{4pt}
\begin{tabular}{@{}l >{\raggedright\arraybackslash}p{0.26\linewidth} >{\raggedright\arraybackslash}p{0.40\linewidth} p{0.12\linewidth}@{}}
\toprule
Domain & Candidate cause & What the data show & Separating stream \\
\midrule
Thermal & schedule change & scheduled setpoint and setpoint in effect move together & SpC \\
Thermal & local override or controller offset & setpoint in effect moves, scheduled setpoint stays & SpC \\
Thermal & terminal-unit fault (stuck damper or valve) & the command moves or saturates while the position stays & ActC, ActP \\
Thermal & local load change & command and position move together in the correcting direction & ActC, ActP \\
Thermal & supply-side capacity loss & supply air temperature off its normal value, peer zones deviate together & SAT, T (peers) \\
Thermal & zone sensor fault & reading departs from peers with setpoints, command and position unchanged & T (peers), ActC, ActP \\
Equipment & commanded change & run command or status changes with power & Run \\
Equipment & load change & flow or delivered output changes with power & Load \\
Equipment & degradation or fault & power rises at unchanged command and load & Run, Load \\
Energy & weather-driven change & use moves with outside air temperature, as on comparable days & OAT \\
Energy & which end use changed & two metered end-use groups fix the third as the residual & Sub2 \\
Air quality & occupancy above design & occupancy rises with CO$_2$ & Occ \\
Air quality & insufficient outdoor air & outdoor-air flow low at normal occupancy & OAF \\
\bottomrule
\end{tabular}
\end{table}

Table~\ref{tab:brickmap} gives the rules that read the stream classes of a real Brick KG
(Section~\ref{sec:real}). A class listed as a root counts wherever the point is, and its Brick
subclasses count with it. A generic class counts only when the point is attached as stated. A
point's attachment is the entity it is a point of: a zone or room, a terminal unit, central HVAC
equipment, or the building. A point on a damper or valve also takes the attachment of the unit that
owns the damper or valve. In the linked census only points that carry a series reference are read. A
point typed with a class that Brick names as the successor of a listed class counts as the listed
class. A metered point attached to nothing and linked to nothing counts as no stream class.

{\footnotesize\linespread{1}\selectfont\setlength{\tabcolsep}{4pt}
\begin{longtable}{@{}l>{\raggedright\arraybackslash}p{0.44\linewidth}>{\raggedright\arraybackslash}p{0.38\linewidth}@{}}
\caption{Brick classes read as canon stream classes in the real-building census.}\label{tab:brickmap}\\
\toprule
Stream & Brick classes & Condition \\
\midrule
\endfirsthead
\toprule
Stream & Brick classes & Condition \\
\midrule
\endhead
\bottomrule
\endlastfoot
T & Zone\_Air\_Temperature\_Sensor, Room\_Air\_Temperature\_Sensor; generic Air\_Temperature\_Sensor, Temperature\_Sensor & generic classes only on a zone or terminal unit \\
Sp & Zone and room air temperature setpoints; effective, occupied and unoccupied heating and cooling setpoints; generic Cooling, Heating and Temperature\_Setpoint & generic classes only on a zone or terminal unit \\
SpC & as Sp & a KG that holds both a scheduled (occupied or unoccupied) and an effective setpoint \\
ActC & Damper\_Position\_Command and \_Setpoint; Valve\_Command; Heating\_Command; Cooling\_Command & on a zone or terminal unit \\
ActP & Damper\_Position\_Sensor; Valve\_Position\_Sensor; generic Position\_Sensor and supply, discharge and air flow sensors & on a zone or terminal unit \\
SAT & Supply\_Air\_Temperature\_Sensor, Discharge\_Air\_Temperature\_Sensor & anywhere \\
OAT & Outside\_Air\_Temperature\_Sensor & anywhere \\
Occ & Occupancy\_Sensor, Occupancy\_Count\_Sensor, Motion\_Sensor & anywhere \\
CO2 & CO2\_Sensor & anywhere \\
OAF & Outside\_Air\_Flow\_Sensor & anywhere \\
Run & on-off, start-stop, enable, run, system and speed statuses and commands; frequency, pump and fan commands; Output\_Frequency\_Sensor & on central HVAC equipment \\
Load & Air\_Flow\_Sensor, Water\_Flow\_Sensor, Thermal\_Power\_Sensor & on central HVAC equipment \\
Whole & electrical meter, or an electric power or energy sensor & on the building or its building meter \\
Pwr & electric power or energy sensor & on central HVAC equipment or a terminal unit \\
Lgt & electric power or energy sensor; illuminance and luminance points & on lighting equipment, or (power) on a zone with no other attachment \\
Sub, Sub2 & a submeter & its end uses, found by following \texttt{feeds} for two hops, give one (Sub) or two (Sub2) end-use groups \\
\end{longtable}}

Table~\ref{tab:cells} applies the requirement table to the four profiles of the sweep, cell by cell.

{\footnotesize\linespread{1}\selectfont\setlength{\tabcolsep}{5pt}
\begin{longtable}{@{}llrrrrr@{}}
\caption{Achieved level of every canon cell at each profile of the sweep (large office, Tampa;
profiles as in Table~\ref{tab:profiles}). The column sums are the numerators of $\CI$ in
Table~\ref{tab:sweep}.}\label{tab:cells}\\
\toprule
Domain & Intent & Target & minimal & comfort & standard & metered \\
\midrule
\endfirsthead
\toprule
Domain & Intent & Target & minimal & comfort & standard & metered \\
\midrule
\endhead
\bottomrule
\endlastfoot
Thermal & status & 2 & 1 & 2 & 2 & 2 \\
Thermal & summary & 2 & 1 & 2 & 2 & 2 \\
Thermal & comparison & 3 & 2 & 2 & 2 & 2 \\
Thermal & verification & 3 & 1 & 2 & 2 & 2 \\
Thermal & diagnosis & 3 & 1 & 2 & 2 & 2 \\
Thermal & explanation & 3 & 1 & 2 & 2 & 2 \\
Energy & status & 1 & 0 & 0 & 0 & 1 \\
Energy & summary & 2 & 0 & 0 & 0 & 2 \\
Energy & comparison & 3 & 0 & 0 & 0 & 2 \\
Energy & diagnosis & 3 & 0 & 0 & 0 & 2 \\
Energy & explanation & 3 & 0 & 0 & 0 & 2 \\
Air quality & status & 2 & 0 & 0 & 0 & 0 \\
Air quality & summary & 2 & 0 & 0 & 0 & 0 \\
Air quality & comparison & 3 & 0 & 0 & 0 & 0 \\
Air quality & verification & 3 & 0 & 0 & 0 & 0 \\
Air quality & diagnosis & 3 & 0 & 0 & 0 & 0 \\
Air quality & explanation & 3 & 0 & 0 & 0 & 0 \\
Equipment & status & 2 & 0 & 0 & 0 & 2 \\
Equipment & summary & 2 & 0 & 0 & 0 & 2 \\
Equipment & comparison & 3 & 0 & 0 & 0 & 2 \\
Equipment & verification & 3 & 0 & 0 & 0 & 2 \\
Equipment & diagnosis & 3 & 0 & 0 & 0 & 2 \\
Equipment & explanation & 3 & 0 & 0 & 0 & 2 \\
Occupancy & status & 2 & 0 & 0 & 2 & 2 \\
Occupancy & summary & 2 & 0 & 0 & 2 & 2 \\
Occupancy & comparison & 2 & 0 & 0 & 2 & 2 \\
Occupancy & diagnosis & 2 & 0 & 0 & 2 & 2 \\
Lighting & status & 2 & 0 & 0 & 0 & 2 \\
Lighting & summary & 2 & 0 & 0 & 0 & 2 \\
Lighting & comparison & 2 & 0 & 0 & 0 & 2 \\
Lighting & diagnosis & 2 & 0 & 0 & 0 & 2 \\
\midrule
\multicolumn{2}{@{}l}{Credited rungs} & 76 & 7 & 12 & 20 & 49 \\
\multicolumn{2}{@{}l}{$\CI$} & & 0.092 & 0.158 & 0.263 & 0.645 \\
\end{longtable}}

\FloatBarrier
\section{The agent's prompt and tools}
\label{app:agent}
The agent is the system under test (Section~\ref{sec:agent}). The reference model is
\texttt{qwen3.8:27b} in its four-bit build (Q4\_K\_M), served by Ollama 0.33.3, model ID
\texttt{22130167c4c2}, with its context raised to 131,072 tokens. The server's default sampling
applies: temperature 1.0, nucleus sampling at 1.0 and top-k 20, with no fixed seed. The full digests
are recorded in the released code. For each question the agent receives the system prompt of
Listing~\ref{lst:prompt}, the question text with its answer shape, and access to the three tools of
Table~\ref{tab:tools}. The agent's code records every turn, the token count of every prompt, and the
outcome (answered, abstained, no answer, or error), and archives the transcript. The model writes
out its reasoning before each reply, and that text is neither archived nor passed to later turns.
The largest prompt of any turn was 62,031 tokens, under half of the window, and no turn of any
model in the study reached its model's window.

\begin{table}[!htbp]
\centering
\caption{The three tools, as declared to the model. In Section~\ref{sec:agent} they are the
KG query, Python and answer tools.}
\label{tab:tools}
\small\linespread{1}\selectfont
\setlength{\tabcolsep}{4pt}
\begin{tabular}{@{}>{\raggedright\arraybackslash}p{0.19\linewidth}>{\raggedright\arraybackslash}p{0.22\linewidth}>{\raggedright\arraybackslash}p{0.53\linewidth}@{}}
\toprule
Tool & Input & Description given to the model \\
\midrule
\texttt{query\_graph} & \texttt{sparql} (string) & Run a SPARQL SELECT query over the building's Brick graph. Returns up to 200 rows as JSON. Prefixes brick:, ref:, rdfs: and rdf: are predefined. \\
\texttt{run\_python} & \texttt{code} (string) & Run standard-library Python 3 in an isolated process. GRAPH\_TTL, CATALOG\_CSV and VALUES\_CSV are predefined file paths. Print what you need; stdout and stderr are returned. \\
\texttt{submit\_answer} & \texttt{answer} (object), \texttt{abstain} (boolean, required), \texttt{needed\_streams} (string) & Hand in the final answer. Call exactly once, as the last step. The answer is in the JSON shape the question gives; abstain is true if the available data cannot determine the answer; when abstaining, name the streams that would be needed. \\
\bottomrule
\end{tabular}
\end{table}

\renewcommand{\thelstlisting}{\Alph{section}.\arabic{lstlisting}}
\begin{lstlisting}[caption={The agent's system prompt, verbatim except for line breaks.},label={lst:prompt}]
You answer one operational question about one building, using only its
data.

You have the building's Brick metadata graph and a time-series store.
The store is two CSV files: CATALOG_CSV (timeseries_id, point_uri,
point_class, unit, quantity_kind, building_id) and VALUES_CSV
(timeseries_id, timestamp, value; hourly). A point in the graph links
to its series through ref:hasExternalReference / ref:hasTimeseriesId.
The store holds only the streams this building exposes, so a point in
the graph may have no series.

Work in three steps. Resolve the entities the question names to their
series ids with query_graph. Fetch and compute with run_python, which
runs standard-library Python with the variables GRAPH_TTL, CATALOG_CSV
and VALUES_CSV already defined as file paths. Only the standard
library is installed, and each call is a fresh process, so nothing
carries over between calls. Do the arithmetic in code, not in your
head. Then call submit_answer once, with the answer in the JSON shape
the question gives.

If the available streams cannot determine the answer, do not guess.
Call submit_answer with abstain set to true and name the streams you
would need.
\end{lstlisting}

\FloatBarrier
{\small\linespread{1}\selectfont\setlength{\tabcolsep}{4pt}
\begin{longtable}{@{}r>{\raggedright\arraybackslash}p{0.12\linewidth}>{\raggedright\arraybackslash}p{0.44\linewidth}>{\raggedright\arraybackslash}p{0.32\linewidth}@{}}
\caption{One episode of the reference agent on the sweep building at the metered profile. The first
query uses relation names the KG does not have and returns nothing, and the agent recovers by
finding the zone by its label. The full transcript is in the released code.}\label{tab:episode}\\
\toprule
\multicolumn{4}{@{}p{0.97\linewidth}@{}}{Question (thermal summary, L1): \emph{What was the mean air
temperature in zone Perimeter\_mid\_ZN\_4 on 2018-01-02 (all 24 hourly values), in
\si{\celsius}?} Answer shape \texttt{\{"value": <number>\}}.} \\
\midrule
Call & Tool & What the agent did & Result \\
\midrule
\endfirsthead
\toprule
Call & Tool & What the agent did & Result \\
\midrule
\endhead
1 & KG query & looked for the zone's temperature point with relation names the KG does not use & no
rows \\
2 & KG query & found the zone by its label & Zone\_F3\_Z05 \\
3 & Python & printed the first lines of the series catalogue & column names and 19 series \\
4 & Python & filtered the catalogue to Zone\_F3\_Z05 & the zone's temperature series and its id \\
5 & Python & read that series for 2 January and averaged the 24 values & 24 values, mean 24.2313 \\
6 & Answer & submitted \texttt{\{"value": 24.2313\}} & ground truth 24.2313, within
$\pm$0.5\,\si{\celsius}, correct \\
\bottomrule
\end{longtable}}

\FloatBarrier
\section{The question bank}
\label{app:bank}
Table~\ref{tab:bank} lists the 64 question classes (Section~\ref{sec:bank}), ordered by domain,
intent and level. Each class requires the stream classes shown, in the abbreviations of
\ref{app:canon}. A class may compute with fewer of those streams, as the weather class uses only the
meter and outside air temperature. Instances bind a zone, an item of equipment, an hour, a date or a
month. Hops count the KG relations an instance must traverse beyond the entity it names. A numeric
answer is correct when it lies within the absolute tolerance, in the question's units, or within the
relative tolerance of the ground-truth value, whichever is met. The five classes marked with a
dagger reproduce analyses from the Mortar application library. The score is the mean over the three
repeats of the sweep, with its range when the repeats differ. A dash marks a class the published
corpus cannot pose. The questions are paraphrased in the table. The verbatim templates, with the
conventions each states, are in the released code. The released code also lists one instance of
every class the published corpus poses, with its ground truth. Four classes span two KG hops, and
three of them instantiate on the published corpus. The two-hop class on the warmest zone under a
named air handler scored 0.78 at the minimal profile, 0.89 at the standard profile and 1.00 at the
comfort and metered profiles, and the other two two-hop classes scored 1.00 at the metered profile.
Of the five classes that reproduce Mortar analyses, only weekday mean energy can be posed on the
published corpus, and it scored 1.00 in every repeat. The other four need streams that neither
source provides. The two energy classes need both outside air temperature, which the published
corpus lacks, and a whole-building meter, which the eight real buildings lack. The airflow class
needs a usable airflow setpoint, and the class on simultaneous heating and cooling needs run status,
load and valve commands. Only one class, the end use behind an increase in energy use, asks for the
attribution directly. An instance is skipped when its conditions are not met, such as two zones too
close in temperature to rank, so a class can yield fewer than three. Where the KG gives one label to
two entities, such as a room and a thermal zone, the question names the entity type. The interval on
realized readiness is a bootstrap that resamples instances within each rung and keeps all repeats of
an instance together, with 2,000 resamples, and Fisher's exact test is two-sided.
Table~\ref{tab:worked} gives worked examples. No L3 class instantiates on the published corpus,
which exports none of the streams they need. For example, the full text of e\_explanation\_2 is
``Between \emph{d1} and \emph{d2}, did whole-building electricity use change more, in absolute
joules summed over hours, during the occupied hours 08:00 to 18:00 (both included) or during the
other hours of the day? Answer occupied or unoccupied.''

{\footnotesize\linespread{1}\selectfont\setlength{\tabcolsep}{3pt}
\begin{longtable}{@{}>{\raggedright\arraybackslash}p{0.19\linewidth}>{\raggedright\arraybackslash}p{0.155\linewidth}>{\raggedright\arraybackslash}p{0.10\linewidth}>{\raggedright\arraybackslash}p{0.125\linewidth}>{\raggedright\arraybackslash}p{0.22\linewidth}>{\raggedright\arraybackslash}p{0.11\linewidth}@{}}
\caption{The 64 question classes, with the reference agent's score at the metered profile of the
sweep. Each class is named as in the released code. A double dagger marks a class
that one fixed answer passes in every instance on the sweep building.}\label{tab:bank}\\
\toprule
Name in code & Cell, level & Requires & Answer & Question (paraphrased) & Score \\
\midrule
\endfirsthead
\toprule
Name in code & Cell, level & Requires & Answer & Question (paraphrased) & Score \\
\midrule
\endhead
\bottomrule
\endlastfoot
t\_\allowbreak{}status\_\allowbreak{}1 & thermal / status, L1 & T & number, $\pm$0.5 & air temperature of a zone at an hour & 1.00 \\
t\_\allowbreak{}status\_\allowbreak{}2$^{\ddagger}$ & thermal / status, L2 & T, Sp & number, $\pm$0.5 & zone temperature minus its cooling setpoint at an hour & 1.00 \\
t\_\allowbreak{}summary\_\allowbreak{}1 & thermal / summary, L1 & T & number, $\pm$0.5 & mean zone temperature over a day & 1.00 \\
t\_\allowbreak{}summary\_\allowbreak{}2$^{\ddagger}$ & thermal / summary, L2 & T, Sp & number, $\pm$1 & hours in a day outside the zone's own setpoint band & 1.00 \\
t\_\allowbreak{}comparison\_\allowbreak{}1 & thermal / comparison, L1 & T & yes/no & was one zone warmer than another at an hour & 1.00 \\
t\_\allowbreak{}comparison\_\allowbreak{}2 & thermal / comparison, L2 & T & number, $\pm$0.5 & temperature of one zone minus another at an hour & 0.89 (0.67--1.00) \\
t\_\allowbreak{}comparison\_\allowbreak{}2hop & thermal / comparison, L2 & T & entity & warmest zone served by a named air handler at an hour (2 hops) & 1.00 \\
t\_\allowbreak{}comparison\_\allowbreak{}3 & thermal / comparison, L3 & T, Sp, ActC, ActP & number, $\pm$0.01 or $\pm$5\% & mean damper command of one zone minus another, 08:00--18:00 & -- \\
t\_\allowbreak{}verification\_\allowbreak{}1 & thermal / verification, L1 & T & yes/no & did a zone stay at or below a stated temperature at every hour of a day & 0.78 (0.67--1.00) \\
t\_\allowbreak{}verification\_\allowbreak{}2 & thermal / verification, L2 & T, Sp & yes/no & did the zone stay within its setpoints at every hour of a day & 1.00 \\
t\_\allowbreak{}verification\_\allowbreak{}3 & thermal / verification, L3 & T, Sp, ActC, ActP & number, $\pm$0.01 or $\pm$5\% & mean damper command over the hours above the cooling setpoint & -- \\
t\_\allowbreak{}verification\_\allowbreak{}3\_\allowbreak{}flow$^{\dagger}$ & thermal / verification, L3 & T, Sp, ActC, ActP (+ airflow setpoint) & number, $\pm$1 & hours the terminal airflow was more than 10\% off its setpoint & -- \\
t\_\allowbreak{}diagnosis\_\allowbreak{}1$^{\ddagger}$ & thermal / diagnosis, L1 & T & entity & warmest zone in the building at an hour & 1.00 \\
t\_\allowbreak{}diagnosis\_\allowbreak{}2$^{\ddagger}$ & thermal / diagnosis, L2 & T, Sp & entity & zone farthest above its own cooling setpoint at an hour & 0.83 (0.50--1.00) \\
t\_\allowbreak{}explanation\_\allowbreak{}1 & thermal / explanation, L1 & T & number, $\pm$0.5 & change in a zone's daily mean temperature between two days & 1.00 \\
t\_\allowbreak{}explanation\_\allowbreak{}2$^{\ddagger}$ & thermal / explanation, L2 & T, Sp & entity & did a zone's mean deviation from its cooling setpoint, 08:00--18:00, increase or decrease between two days & 1.00 \\
e\_\allowbreak{}status\_\allowbreak{}1 & energy / status, L1 & Whole & number, $\pm$2\% & whole-building electricity on a day & 1.00 \\
e\_\allowbreak{}summary\_\allowbreak{}1 & energy / summary, L1 & Whole & number, $\pm$2\% & highest hourly whole-building use in a month & 1.00 \\
e\_\allowbreak{}summary\_\allowbreak{}1\_\allowbreak{}weekday$^{\dagger}$ & energy / summary, L1 & Whole & number, $\pm$2\% & mean weekday daily use in a month & 1.00 \\
e\_\allowbreak{}summary\_\allowbreak{}2 & energy / summary, L2 & Whole, Sub & number, $\pm$0.02 & share of a day's electricity that went to lighting & 1.00 \\
e\_\allowbreak{}comparison\_\allowbreak{}1 & energy / comparison, L1 & Whole & yes/no & did the building use more electricity on one day than on another & 1.00 \\
e\_\allowbreak{}comparison\_\allowbreak{}2 & energy / comparison, L2 & Whole & number, $\pm$2\% & whole-building use on one day minus another & 1.00 \\
e\_\allowbreak{}comparison\_\allowbreak{}3 & energy / comparison, L3 & Whole, Sub2, OAT & entity & end use with the largest increase between two days (2 hops) & -- \\
e\_\allowbreak{}comparison\_\allowbreak{}3\_\allowbreak{}weather$^{\dagger}$ & energy / comparison, L3 & Whole, Sub2, OAT & number, $\pm$5\% & change in daily use per degree of outside temperature & -- \\
e\_\allowbreak{}diagnosis\_\allowbreak{}1 & energy / diagnosis, L1 & Whole & entity & date of the highest daily use in a month & 1.00 \\
e\_\allowbreak{}diagnosis\_\allowbreak{}2 & energy / diagnosis, L2 & Whole & number, $\pm$1 & days in a month more than 10\% above the month's mean daily use & 1.00 \\
e\_\allowbreak{}explanation\_\allowbreak{}1 & energy / explanation, L1 & Whole & number, $\pm$0.01 or $\pm$5\% & fractional change in daily use between two days & 1.00 \\
e\_\allowbreak{}explanation\_\allowbreak{}2 & energy / explanation, L2 & Whole & entity & did use change more in the occupied hours or the other hours between two days & 0.67 \\
e\_\allowbreak{}explanation\_\allowbreak{}occ$^{\dagger}$ & energy / explanation, L3 & Whole, Sub2, OAT, Occ & number, $\pm$0.05 & correlation of hourly use with total occupancy on a day & -- \\
q\_\allowbreak{}status\_\allowbreak{}1 & equipment / status, L1 & Pwr & number, $\pm$1 or $\pm$2\% & power of a named item of equipment at an hour & 1.00 \\
q\_\allowbreak{}status\_\allowbreak{}2 & equipment / status, L2 & Pwr & number, $\pm$0.02 & power at an hour as a fraction of the equipment's own peak & 1.00 \\
q\_\allowbreak{}summary\_\allowbreak{}1 & equipment / summary, L1 & Pwr & number, $\pm$1 or $\pm$2\% & mean power of an item over a day & 1.00 \\
q\_\allowbreak{}summary\_\allowbreak{}2 & equipment / summary, L2 & Pwr & number, $\pm$1 & hours in a day an item ran above half its own peak & 1.00 \\
q\_\allowbreak{}comparison\_\allowbreak{}1 & equipment / comparison, L1 & Pwr & yes/no & did one item draw more power than another at an hour & 0.89 (0.67--1.00) \\
q\_\allowbreak{}comparison\_\allowbreak{}2 & equipment / comparison, L2 & Pwr & entity & which of three items used the most electricity on a day (2 hops) & 1.00 \\
q\_\allowbreak{}comparison\_\allowbreak{}3 & equipment / comparison, L3 & Pwr, Run, Load & number, $\pm$5\% & mean air flow of one item minus another on a day & -- \\
q\_\allowbreak{}verification\_\allowbreak{}1 & equipment / verification, L1 & Pwr & yes/no & did an item exceed a stated power at any hour of a day & 1.00 \\
q\_\allowbreak{}verification\_\allowbreak{}2 & equipment / verification, L2 & Pwr & yes/no & did an item stay below 90\% of its own peak at every hour of a day & 0.89 (0.67--1.00) \\
q\_\allowbreak{}verification\_\allowbreak{}3 & equipment / verification, L3 & Pwr, Run, Load & number, $\pm$1 & hours an item drew power while its run signal was off & -- \\
q\_\allowbreak{}verification\_\allowbreak{}3\_\allowbreak{}simul$^{\dagger}$ & equipment / verification, L3 & Pwr, Run, Load (+ valve commands) & number, $\pm$1 & hours heating and cooling valves were both open & -- \\
q\_\allowbreak{}diagnosis\_\allowbreak{}1$^{\ddagger}$ & equipment / diagnosis, L1 & Pwr & entity & item drawing the most power at an hour & 1.00 \\
q\_\allowbreak{}diagnosis\_\allowbreak{}2 & equipment / diagnosis, L2 & Pwr & entity & item that ran closest to its own peak on a day & 1.00 \\
q\_\allowbreak{}explanation\_\allowbreak{}1 & equipment / explanation, L1 & Pwr & number, $\pm$5\% & change in an item's daily energy between two days & 1.00 \\
q\_\allowbreak{}explanation\_\allowbreak{}2 & equipment / explanation, L2 & Pwr & entity & did an item's use change more in the occupied hours or the other hours between two days & 0.67 (0.33--1.00) \\
a\_\allowbreak{}status\_\allowbreak{}1 & air quality / status, L1 & CO2 & number, $\pm$2\% & CO$_2$ in a zone at an hour & -- \\
a\_\allowbreak{}summary\_\allowbreak{}2 & air quality / summary, L2 & CO2, Occ & number, $\pm$1 & occupied hours with CO$_2$ above 800 ppm on a day & -- \\
a\_\allowbreak{}comparison\_\allowbreak{}2 & air quality / comparison, L2 & CO2 & number, $\pm$5 or $\pm$5\% & mean CO$_2$ of one zone minus another on a day & -- \\
a\_\allowbreak{}verification\_\allowbreak{}3 & air quality / verification, L3 & CO2, Occ, OAF & yes/no & was outdoor-air flow higher when occupied than when not & -- \\
o\_\allowbreak{}status\_\allowbreak{}1 & occupancy / status, L1 & Occ & number, $\pm$1 or $\pm$5\% & people in a zone at an hour & 1.00 \\
o\_\allowbreak{}status\_\allowbreak{}2 & occupancy / status, L2 & Occ & yes/no & were more people in a zone at an hour than its mean for 08:00--18:00 & 0.89 (0.67--1.00) \\
o\_\allowbreak{}summary\_\allowbreak{}1 & occupancy / summary, L1 & Occ & number, $\pm$1 or $\pm$5\% & highest hourly occupancy of a zone on a day & 1.00 \\
o\_\allowbreak{}summary\_\allowbreak{}2 & occupancy / summary, L2 & Occ & number, $\pm$1 or $\pm$5\% & mean occupancy 08:00--18:00 on a day & 1.00 \\
o\_\allowbreak{}comparison\_\allowbreak{}1 & occupancy / comparison, L1 & Occ & yes/no & were more people in one zone than another at an hour & 1.00 \\
o\_\allowbreak{}comparison\_\allowbreak{}2 & occupancy / comparison, L2 & Occ & number, $\pm$1 or $\pm$5\% & mean occupancy of one zone minus another, 08:00--18:00 & 0.89 (0.67--1.00) \\
o\_\allowbreak{}diagnosis\_\allowbreak{}1 & occupancy / diagnosis, L1 & Occ & number, exact & hour of the day with the most people in a zone & 1.00 \\
o\_\allowbreak{}diagnosis\_\allowbreak{}2 & occupancy / diagnosis, L2 & Occ & number, $\pm$1 & hours occupied outside 08:00--18:00 on a day & 1.00 \\
l\_\allowbreak{}status\_\allowbreak{}1 & lighting / status, L1 & Lgt & number, $\pm$1 or $\pm$2\% & lighting power of a zone at an hour & 1.00 \\
l\_\allowbreak{}status\_\allowbreak{}2 & lighting / status, L2 & Lgt, Occ & yes/no & lights on while the zone was unoccupied at an hour & 1.00 \\
l\_\allowbreak{}summary\_\allowbreak{}1 & lighting / summary, L1 & Lgt & number, $\pm$1 or $\pm$2\% & mean lighting power of a zone over a day & 1.00 \\
l\_\allowbreak{}summary\_\allowbreak{}2 & lighting / summary, L2 & Lgt & number, $\pm$2\% & lighting energy of a zone on a day & 1.00 \\
l\_\allowbreak{}comparison\_\allowbreak{}1 & lighting / comparison, L1 & Lgt & yes/no & was one zone's lighting power higher than another's at an hour & 0.89 (0.67--1.00) \\
l\_\allowbreak{}comparison\_\allowbreak{}2 & lighting / comparison, L2 & Lgt & entity & zone with the most lighting energy on a day (2 hops) & 1.00 \\
l\_\allowbreak{}diagnosis\_\allowbreak{}1$^{\ddagger}$ & lighting / diagnosis, L1 & Lgt & entity & zone with the highest lighting power at an hour & 1.00 \\
l\_\allowbreak{}diagnosis\_\allowbreak{}2 & lighting / diagnosis, L2 & Lgt, Occ & number, $\pm$1 & hours lights were on while the zone was unoccupied & 0.89 (0.67--1.00) \\
\end{longtable}}

{\footnotesize\linespread{1}\selectfont\setlength{\tabcolsep}{3pt}
\begin{longtable}{@{}>{\raggedright\arraybackslash}p{0.21\linewidth}>{\raggedright\arraybackslash}p{0.30\linewidth}>{\raggedright\arraybackslash}p{0.27\linewidth}>{\raggedright\arraybackslash}p{0.16\linewidth}@{}}
\caption{Worked examples on the sweep building at the metered profile, one per answer type and
domain, and a typical miss. The ground truth of the instance shown is followed, in brackets, by
those of the class's other two instances. The agent's answers are its three
repeats.}\label{tab:worked}\\
\toprule
Example & Question (abridged) & Ground truth (other instances) & Agent, 3 repeats \\
\midrule
\endfirsthead
\toprule
Example & Question (abridged) & Ground truth (other instances) & Agent, 3 repeats \\
\midrule
\endhead
L1 number\newline(t\_status\_1) & air temperature of DataCenter\_\allowbreak{}basement\_\allowbreak{}ZN\_\allowbreak{}6 at 2018-01-02 08:00,
\si{\celsius} & 27.0 (27.001, 24.0) & 27 $\times$3 \\ \addlinespace[2pt]
L2 difference\newline(t\_comparison\_2) & Basement minus the HVAC zone DataCenter\_\allowbreak{}top\_\allowbreak{}ZN\_\allowbreak{}6 at 2018-01-14
16:00, \si{\celsius} & $-3.0$ ($-0.899$, $-3.001$) & $-3$ $\times$3 \\ \addlinespace[2pt]
yes/no\newline(t\_verification\_2) & did Perimeter\_\allowbreak{}mid\_\allowbreak{}ZN\_\allowbreak{}4 stay within its setpoints all of 2018-07-04
& false (true, true) & false $\times$3 \\ \addlinespace[2pt]
entity, lighting\newline(l\_comparison\_2) & which of three zones used the most lighting energy on
2018-10-12 & Core\_\allowbreak{}bottom (Core\_\allowbreak{}bottom, Core\_\allowbreak{}mid) & Core\_\allowbreak{}bottom $\times$3 \\ \addlinespace[2pt]
two KG hops\newline(t\_comparison\_\allowbreak{}2hop) & warmest zone served by AHU02 at 2018-10-09 11:00 &
Perimeter\_\allowbreak{}mid\_\allowbreak{}ZN\_\allowbreak{}2 (Perimeter\_\allowbreak{}top\_\allowbreak{}ZN\_\allowbreak{}1, Perimeter\_\allowbreak{}mid\_\allowbreak{}ZN\_\allowbreak{}2) & Zone\_\allowbreak{}F3\_\allowbreak{}Z03 (its identifier) $\times$3 \\ \addlinespace[2pt]
energy\newline(e\_summary\_1) & highest hourly whole-building electricity use in 2018-10, J &
$5.22\times10^{9}$ ($5.82\times10^{9}$, $4.29\times10^{9}$) & $5.22\times10^{9}$ $\times$3 \\ \addlinespace[2pt]
equipment\newline(q\_summary\_1) & mean power of Chiller01 on 2018-10-10, W & 85,849 (3,358; 7,135) &
85,849 $\times$3 \\ \addlinespace[2pt]
occupancy\newline(o\_status\_1) & people in Perimeter\_\allowbreak{}top\_\allowbreak{}ZN\_\allowbreak{}2 at 2018-10-11 09:00 & 10.3 (1.85, 142.1)
& 10.3 $\times$3 \\ \addlinespace[2pt]
a miss\newline(t\_verification\_1) & did Perimeter\_\allowbreak{}bot\_\allowbreak{}ZN\_\allowbreak{}4 stay at or below 26\,\si{\celsius} all of
2018-10-25 & false (false, true) & abstained, false, abstained \\
\bottomrule
\end{longtable}}

\FloatBarrier
\section{Diagnosis scenarios}
\label{app:diag}
In each scenario question (Section~\ref{sec:faults}) the answer shape is \texttt{\{"cause": "<one of
the candidate causes, copied exactly>"\}}, and the agent may abstain as in every other episode. The
question reads: ``\{observation\} Which one of these causes do the data support: \{cause A\};
\{cause B\}? Use only the data available for this building. If the available streams cannot tell
these causes apart, abstain and name the streams you would need.'' The two scenarios of a pair share
one observation. For pair~A it is that from mid-June the zone has run about 1\,\si{\celsius} cooler
during occupied hours than before, while the other core zones have not changed. For pair~B it is
that since July the reported temperature of the zone has risen relative to the other zones, followed
by the premise ``The reported zone temperature comes from a monitoring sensor. The zone's controller
uses its own thermostat, which is not in the data.'' Both causes act on one zone, so the unchanged
neighbouring zones that the observation mentions do not separate them. In pair~A, EnergyPlus
emulates a biased thermostat by shifting the setpoint in effect and leaving the schedule unchanged,
so the offset is posed as ``local override or controller offset'', and its rival lowers the zone's
own heating and cooling setpoint schedules from the same date. In pair~B, the drift is added after
simulation to the monitoring series with the telemetry injector of BuildStream, the software that
generated BATS. The scheduled setpoint of the supplement comes from the simulation of the faulted
building, which reproduces the release exactly. The damper position comes from a re-simulation of
the clean building with one added output variable, attached to the damper point the release
declares. The re-simulation matches 65 of the 93 released series and differs on the other 28 by at
most 0.063\,\si{\celsius} in zone temperature, so the three scenarios built on the clean building
use the re-simulated series throughout.

{\footnotesize\linespread{1}\selectfont\setlength{\tabcolsep}{4pt}
\begin{longtable}{@{}>{\raggedright\arraybackslash}p{0.17\linewidth}>{\raggedright\arraybackslash}p{0.42\linewidth}>{\raggedright\arraybackslash}p{0.22\linewidth}>{\raggedright\arraybackslash}p{0.10\linewidth}@{}}
\caption{The four diagnosis scenarios, all in the large office in Tampa. Base streams are T and Sp
in every scenario. The separating streams come from the supplement to the release.}\label{tab:scenarios}\\
\toprule
Scenario & Injection & True cause / rival & \hspace{0pt}Separating stream \\
\midrule
\endfirsthead
\toprule
Scenario & Injection & True cause / rival & \hspace{0pt}Separating stream \\
\midrule
\endhead
\bottomrule
\endlastfoot
A: \hspace{0pt}thermostat offset & EnergyPlus fault model, 1\,\si{\celsius} offset in the Core\_bottom zone of the published faulted building from 15 June & local override or controller offset / schedule change & SpC \\
A: schedule change & the zone's own heating and cooling setpoint schedules lowered by 1\,\si{\celsius} from 15 June (re-simulated clean building) & schedule change / local override or controller offset & SpC \\
B: sensor drift & BuildStream telemetry injector, drift to $+1$\,\si{\celsius} on the reported temperature of the Core\_mid zone from 1 July to year end (re-simulated clean building) & zone sensor fault / local load change & ActP \\
B: internal-load step & an always-on equipment load of 40\,\si{\kilo\watt} in the Core\_mid zone from 1 July (re-simulated clean building) & local load change / zone sensor fault & ActP \\
\end{longtable}}

Before the runs, each pair was checked against the clean building over occupied weekday hours, from
16 to 29 June for pair~A and from October to December for pair~B (Table~\ref{tab:pairs}). Within a
pair the base streams move alike and the peer zone does not, so the causes are not identifiable in
the base condition. The separating stream differs clearly between the two scenarios of a pair, so
they are identifiable in the separating condition.

{\footnotesize\linespread{1}\selectfont\setlength{\tabcolsep}{4pt}
\begin{longtable}{@{}llrrrr@{}}
\caption{Look-alike and separation checks: change against the clean building, mean over occupied
weekday hours. The peer is the core zone on another floor.}\label{tab:pairs}\\
\toprule
Pair & Scenario & Zone T & Setpoint in effect & Peer T & Separating stream \\
\midrule
\endfirsthead
\toprule
Pair & Scenario & Zone T & Setpoint in effect & Peer T & Separating stream \\
\midrule
\endhead
\bottomrule
\endlastfoot
A & thermostat offset & $-0.87$\,\si{\celsius} & $-1.00$\,\si{\celsius} & $+0.01$\,\si{\celsius} & SpC $+0.00$\,\si{\celsius} \\
A & schedule change & $-0.91$\,\si{\celsius} & $-1.00$\,\si{\celsius} & $+0.01$\,\si{\celsius} & SpC $-1.00$\,\si{\celsius} \\
B & sensor drift & $+0.75$\,\si{\celsius} & $0.00$\,\si{\celsius} & $0.00$\,\si{\celsius} & damper $+0.000$ \\
B & internal-load step & $+0.67$\,\si{\celsius} & $0.00$\,\si{\celsius} & $0.00$\,\si{\celsius} & damper $+0.180$ \\
\end{longtable}}

\FloatBarrier
\section{The real buildings: series processing and census checks}
\label{app:realdata}
The series of the eight buildings run in depth (Section~\ref{sec:realq}) are converted to hourly
means on the UTC clock of Section~\ref{sec:realq}. No KG declares a unit on any temperature point,
and an air temperature or setpoint series, the only temperatures the questions use, whose median
exceeds 35 is read as Fahrenheit and converted. A series is usable when it holds a value for at
least 90\% of the year's hours and takes at least three distinct hourly values, which excludes stuck
and dead points. Each declared series is classed as usable, as present but unusable, or as absent
from the release. The rule's sensitivity is tested by holding each value until the next sample, and
by varying the coverage from 80\% to 95\% and the distinct values from two to three
(Section~\ref{sec:res-realq}). The ground-truth values of one instance of each of two classes per
building were recomputed from the raw files with a script independent of the study's code, and all
16 matched.

The revision of the Mortar graphs used for the census retypes its 16 former building meters, four of
them linked, as generic energy sensors with no link to the building. The mapping rules count none of
these points. Counted as whole-building meters, they raise the ceiling of four buildings and the
maximum to 0.513. The strict mapping lowers the maximum to 0.263, because the power points of the
Mortar KGs are generic power sensors that count only through their attachment.

Of the ten Mortar applications that compute something, three can run on some KG: zone comfort
evaluation on 23 of the 37, the comparison of sensors with their setpoints on 16, and the rogue-zone
airflow check on one. The five energy applications run on none, because no KG in the revision used
types a building meter. On the June revision, which still typed them, two energy applications run on
four KGs. The application count and the ceiling disagree most where the ceiling comes from equipment
streams (power, run status and load), which no Mortar application uses.

\bibliographystyle{unsrtnat}
\bibliography{refs}

@article{zhang2025llmagentbuilding,
  title   = {Large language model-based agent schema and library for automated building energy analysis and modeling},
  author  = {Zhang, Liang and Fu, Xiao and Li, Yu and Chen, Jianli},
  journal = {Automation in Construction},
  volume  = {176},
  pages   = {106244},
  year    = {2025},
  doi     = {10.1016/j.autcon.2025.106244}
}

@article{wang2022dtbim,
  title   = {Digital twin-enabled built environment sensing and monitoring through semantic enrichment of {BIM} with {SensorML}},
  author  = {Wang, Tian and Gan, Vincent J. L. and Hu, Daofang and Liu, Haosen},
  journal = {Automation in Construction},
  volume  = {144},
  pages   = {104625},
  year    = {2022},
  doi     = {10.1016/j.autcon.2022.104625}
}

@article{xie2023dtfdd,
  title   = {Digital twin enabled fault detection and diagnosis process for building {HVAC} systems},
  author  = {Xie, Xiangyu and Merino, Jorge and Moretti, Nicola and Pauwels, Pieter and Chang, Ji Youn and Parlikad, Ajith Kumar},
  journal = {Automation in Construction},
  volume  = {146},
  pages   = {104695},
  year    = {2023},
  doi     = {10.1016/j.autcon.2023.104695}
}

@article{ma2025dtbuilding,
  title   = {Digital twin for {3D} interactive building operations: integrating {BIM}, {IoT}-enabled building automation systems, {AI}, and mixed reality},
  author  = {Ma, Tianyou and Xiao, Fu and Zhang, Chong and Zhang, Jing and Zhang, Hanbei and Xu, Kan and Luo, Xiaowei},
  journal = {Automation in Construction},
  volume  = {176},
  pages   = {106277},
  year    = {2025},
  doi     = {10.1016/j.autcon.2025.106277}
}

@techreport{verbeke2020sri,
  title   = {Final report on the technical support to the development of a smart readiness indicator for buildings},
  author  = {Verbeke, Stijn and Aerts, Dorien and Reynders, Glenn and Ma, Yixiao and Waide, Paul},
  institution = {European Commission, Directorate-General for Energy},
  year    = {2020},
  note    = {Study on the Smart Readiness Indicator (SRI)}
}

@misc{ec2020sri,
  title  = {Commission Delegated Regulation ({EU}) 2020/2155 supplementing Directive ({EU}) 2010/31 establishing an optional common scheme for rating the smart readiness of buildings},
  author = {{European Commission}},
  year   = {2020},
  howpublished = {Official Journal of the European Union L 431},
  note = {\url{https://eur-lex.europa.eu/eli/reg_del/2020/2155/oj}}
}

@misc{wiredscore2026,
  title = {{SmartScore} certification},
  author = {{WiredScore}},
  howpublished = {\url{https://wiredscore.com/certify-a-building/smartscore/}},
  note = {accessed September 2026}
}

@misc{ashrae223p,
  title = {{ASHRAE} Standard {223P}: Designation and Classification of Semantic Tags for Building Data},
  author = {{ASHRAE}},
  year = {2023},
  howpublished = {American Society of Heating, Refrigerating and Air-Conditioning Engineers}
}

@inproceedings{balaji2016brick,
  title     = {{Brick}: Towards a Unified Metadata Schema For Buildings},
  author    = {Balaji, Bharathan and Bhattacharya, Arka and Fierro, Gabe and Gao, Jingkun and Gluck, Joshua and Hong, Dezhi and others},
  booktitle = {Proc.\ 3rd ACM Int.\ Conf.\ on Systems for Energy-Efficient Built Environments (BuildSys)},
  year      = {2016},
  doi = {10.1145/2993422.2993577}
}

@article{balaji2018brick,
  title   = {{Brick}: Metadata schema for portable smart building applications},
  author  = {Balaji, Bharathan and Bhattacharya, Arka and Fierro, Gabe and others},
  journal = {Applied Energy},
  volume  = {226},
  pages   = {1273--1292},
  year    = {2018},
  doi     = {10.1016/j.apenergy.2018.02.091}
}

@misc{ashrae90_1_2022,
  title = {{ANSI/ASHRAE/IES} Standard 90.1-2022: Energy Standard for Sites and Buildings Except Low-Rise Residential Buildings},
  author = {{ASHRAE}}, year = {2022}
}

@misc{ashrae_gdl36,
  title = {{ASHRAE} Guideline 36-2021: High-Performance Sequences of Operation for {HVAC} Systems},
  author = {{ASHRAE}}, year = {2021}
}

@misc{lawrence2017drl,
  title  = {Data Readiness Levels},
  author = {Lawrence, Neil D.},
  year   = {2017},
  howpublished = {arXiv:1705.02245}
}

@article{boje2020dt,
  title   = {Towards a semantic Construction Digital Twin: Directions for future research},
  author  = {Boje, Calin and Guerriero, Annie and Kubicki, Sylvain and Rezgui, Yacine},
  journal = {Automation in Construction},
  volume  = {114}, pages = {103179}, year = {2020},
  doi     = {10.1016/j.autcon.2020.103179}
}

@inproceedings{mulayim2025buildingqa,
  title     = {{BuildingQA}: A Benchmark for Natural Language Question Answering over Building Knowledge Graphs},
  author    = {Mulayim, Ozan Baris and Anwar, Avia and Saka, Umut Mete and Paul, Lazlo and Prakash, Anand Krishnan and Fierro, Gabe and Pritoni, Marco and Berg{\'e}s, Mario},
  booktitle = {Proc.\ 12th ACM Int.\ Conf.\ on Systems for Energy-Efficient Buildings, Cities, and Transportation (BuildSys)},
  year      = {2025},
  doi       = {10.1145/3736425.3770097}
}

@inproceedings{jung2026bricktrace,
  title     = {{BrickTrace}: Separating Ontology Reasoning from Building Grounding in Knowledge Graph Question Answering},
  author    = {Jung, Wooyoung},
  booktitle = {Proc.\ 2026 ACM Sustainability Week},
  pages     = {438--442},
  year      = {2026},
  doi       = {10.1145/3765611.3815486}
}

@article{jung2026hema,
  title   = {Multi-agent home energy management assistant ({HEMA})},
  author  = {Jung, Wooyoung},
  journal = {SoftwareX},
  volume  = {34},
  pages   = {102633},
  year    = {2026},
  doi     = {10.1016/j.softx.2026.102633}
}

@article{jung2026cot,
  title   = {Chain-of-Thought Prompting for Generating Personalized Home Energy-Saving Strategies},
  author  = {Jung, Wooyoung},
  journal = {E3S Web of Conferences},
  volume  = {716},
  pages   = {06015},
  year    = {2026},
  doi     = {10.1051/e3sconf/202671606015}
}

@inproceedings{fierro2018mortar,
  title     = {Mortar: An Open Testbed for Portable Building Analytics},
  author    = {Fierro, Gabe and Pritoni, Marco and AbdelBaky, Moustafa and Raftery, Paul and Peffer, Therese and Thomson, Greg and Culler, David E.},
  booktitle = {Proc.\ 5th ACM Int.\ Conf.\ on Systems for Energy-Efficient Built Environments (BuildSys)},
  year      = {2018},
  doi = {10.1145/3276774.3276796}
}

@inproceedings{chen2021finqa,
  title     = {{FinQA}: A Dataset of Numerical Reasoning over Financial Data},
  author    = {Chen, Zhiyu and Chen, Wenhu and Smiley, Charese and Shah, Sameena and others},
  booktitle = {Proc.\ EMNLP},
  year      = {2021},
  doi = {10.18653/v1/2021.emnlp-main.300}
}

@inproceedings{zhu2021tatqa,
  title     = {TAT-{QA}: A Question Answering Benchmark on a Hybrid of Tabular and Textual Content in Finance},
  author    = {Zhu, Fengbin and Lei, Wenqiang and Huang, Youcheng and others},
  booktitle = {Proc.\ ACL},
  year      = {2021},
  doi = {10.18653/v1/2021.acl-long.254}
}

@inproceedings{merrill2024timeseries,
  title     = {Language Models Still Struggle to Zero-shot Reason about Time Series},
  author    = {Merrill, Mike A. and Tan, Mingtian and Gupta, Vinayak and Hartvigsen, Tom and Althoff, Tim},
  booktitle = {Findings of the Association for Computational Linguistics: EMNLP 2024},
  year      = {2024},
  doi = {10.18653/v1/2024.findings-emnlp.201}
}

@article{llmkg2025survey,
  title   = {Large Language Models Meet Knowledge Graphs for Question Answering: Synthesis and Opportunities},
  author  = {Ma, Chuangtao and Chen, Yongrui and Wu, Tianxing and Khan, Arijit and Wang, Haofen},
  journal = {arXiv:2505.20099},
  year    = {2025}
}

@article{granderson2020data,
  title   = {Building fault detection data to aid diagnostic algorithm creation and performance testing},
  author  = {Granderson, Jessica and Lin, Guanjing and Harding, Aaron and Im, Piljae and Chen, Yan},
  journal = {Scientific Data},
  volume  = {7}, number = {1}, pages = {65},
  year    = {2020},
  doi = {10.1038/s41597-020-0398-6}
}

@article{frank2019framework,
  title   = {A performance evaluation framework for building fault detection and diagnosis algorithms},
  author  = {Frank, Stephen and Lin, Guanjing and Jin, Xin and Singla, Rupam and Farthing, Amanda and Granderson, Jessica},
  journal = {Energy and Buildings},
  volume  = {192}, pages = {84--92}, year = {2019},
  doi     = {10.1016/j.enbuild.2019.03.024}
}

@inproceedings{johnson2017clevr,
  title     = {{CLEVR}: A Diagnostic Dataset for Compositional Language and Elementary Visual Reasoning},
  author    = {Johnson, Justin and Hariharan, Bharath and van der Maaten, Laurens and Fei-Fei, Li and Zitnick, C. Lawrence and Girshick, Ross},
  booktitle = {Proc.\ IEEE Conf.\ on Computer Vision and Pattern Recognition (CVPR)},
  year      = {2017},
  doi = {10.1109/CVPR.2017.215}
}

@article{geirhos2020shortcut,
  title   = {Shortcut learning in deep neural networks},
  author  = {Geirhos, Robert and Jacobsen, J{\"o}rn-Henrik and Michaelis, Claudio and others},
  journal = {Nature Machine Intelligence},
  volume  = {2}, pages = {665--673}, year = {2020},
  doi = {10.1038/s42256-020-00257-z}
}

@article{crawley2001energyplus,
  title   = {{EnergyPlus}: creating a new-generation building energy simulation program},
  author  = {Crawley, Drury B. and Lawrie, Linda K. and Winkelmann, Frederick C. and others},
  journal = {Energy and Buildings},
  volume  = {33}, number = {4}, pages = {319--331}, year = {2001},
  doi     = {10.1016/S0378-7788(00)00114-6}
}

@techreport{doe_prototypes,
  title  = {Enhancements to {ASHRAE} Standard 90.1 Prototype Building Models},
  author = {Goel, Supriya and Athalye, Rahul A. and Wang, Weimin and others},
  institution = {Pacific Northwest National Laboratory (PNNL-23269)}, year = {2014},
  doi = {10.2172/1129366}
}

@article{zhang2017faultmodel,
  title   = {Modeling of {HVAC} operational faults in building performance simulation},
  author  = {Zhang, Rongpeng and Hong, Tianzhen},
  journal = {Applied Energy}, volume = {202}, pages = {178--188}, year = {2017},
  doi = {10.1016/j.apenergy.2017.05.153}
}

@misc{lee2025hvacqa,
  title  = {{LLM}-based Question-Answer Framework for Sensor-driven {HVAC} System Interaction},
  author = {Lee, Sungmin and Kang, Minju and Lee, Joonhee and Lee, Seungyong and Kim, Dongju
            and Hong, Jingi and Shin, Jun and Zhang, Pei and Ko, JeongGil},
  year   = {2025},
  howpublished = {arXiv:2507.04748}
}

@article{zhang2025hvacfdd,
  title   = {Domain-specific large language models for fault diagnosis of heating, ventilation, and air conditioning systems by labeled-data-supervised fine-tuning},
  author  = {Zhang, Jian and Zhang, Chaobo and Lu, Jie and Zhao, Yang},
  journal = {Applied Energy},
  volume  = {377},
  pages   = {124378},
  year    = {2025},
  doi     = {10.1016/j.apenergy.2024.124378}
}

@misc{song2023industrialcontrol,
  title  = {Pre-Trained Large Language Models for Industrial Control},
  author = {Song, Lei and Zhang, Chuheng and Zhao, Li and Bian, Jiang},
  year   = {2023},
  howpublished = {arXiv:2308.03028}
}

@article{liu2025llmbuilding,
  title   = {Large language models for building energy applications: Opportunities and challenges},
  author  = {Liu, Mingzhe and Zhang, Liang and Chen, Jianli and Chen, Wei-An and Yang, Zhiyao
            and Lo, L. James and Wen, Jin and O'Neill, Zheng},
  journal = {Building Simulation},
  volume  = {18},
  number  = {2},
  pages   = {225--234},
  year    = {2025},
  doi     = {10.1007/s12273-025-1235-9}
}

@article{arslan2026llmsurvey,
  title   = {Large language models in building energy applications: A survey},
  author  = {Arslan, Muhammad and Munawar, Saba},
  journal = {Energy and Buildings},
  volume  = {352},
  pages   = {116800},
  year    = {2026},
  doi     = {10.1016/j.enbuild.2025.116800}
}

@article{granderson2023labeled,
  title   = {A labeled dataset for building {HVAC} systems operating in faulted and fault-free states},
  author  = {Granderson, Jessica and Lin, Guanjing and Chen, Yimin and Casillas, Armando and Wen, Jin and Chen, Zhelun and Im, Piljae and Huang, Sen and Ling, Jiazhen},
  journal = {Scientific Data},
  volume  = {10},
  number  = {1},
  year    = {2023},
  doi     = {10.1038/s41597-023-02197-w}
}

@incollection{saref2023etsi,
  title   = {The {ETSI SAREF} Ontology for Smart Applications: A Long Path of Development and Evolution},
  author  = {García‐Castro, Raúl and Lefrançois, Maxime and Poveda‐Villalón, María and Daniele, Laura},
  booktitle= {Energy Smart Appliances},
  pages   = {183-215},
  year    = {2023},
  doi     = {10.1002/9781119899457.ch7}
}

@inproceedings{avila2025kgqa,
  title   = {A Framework for Question Answering on Knowledge Graphs Using Large Language Models},
  author  = {Avila, Caio Viktor S. and Casanova, Marco A. and Vidal, Vânia M. P.},
  booktitle= {Lecture Notes in Computer Science},
  pages   = {168-172},
  year    = {2025},
  doi     = {10.1007/978-3-031-78952-6_20}
}

@article{li2022virtualsensor,
  title   = {Validation of virtual sensor-assisted {Bayesian} inference-based in-situ sensor calibration strategy for building {HVAC} systems},
  author  = {Li, Guannan and Xiong, Jiahao and Sun, Shaobo and Chen, Jian},
  journal = {Building Simulation},
  volume  = {16},
  number  = {2},
  pages   = {185-203},
  year    = {2022},
  doi     = {10.1007/s12273-022-0935-7}
}

@article{billanes2025smartbuilding,
  title   = {Data-Driven Technologies for Energy Optimization in Smart Buildings: A Scoping Review},
  author  = {Billanes, Joy Dalmacio and Ma, Zheng Grace and Jørgensen, Bo Nørregaard},
  journal = {Energies},
  volume  = {18},
  number  = {2},
  pages   = {290},
  year    = {2025},
  doi     = {10.3390/en18020290}
}

@article{jobe2026intent,
  title   = {Lightweight large language models for intent recognition and action invocation in building energy management},
  author  = {Liu, Deli and Zhou, Xiaoping and Li, Yu},
  journal = {Journal of Building Engineering},
  volume  = {125},
  pages   = {116170},
  year    = {2026},
  doi     = {10.1016/j.jobe.2026.116170}
}

@article{zhang2024llmbem,
  title   = {Advancing building energy modeling with large language models: Exploration and case studies},
  author  = {Zhang, Liang and Chen, Zhelun and Ford, Vitaly},
  journal = {Energy and Buildings},
  volume  = {323},
  pages   = {114788},
  year    = {2024},
  doi     = {10.1016/j.enbuild.2024.114788}
}

@article{parn2017bimfm,
  title   = {The building information modelling trajectory in facilities management: A review},
  author  = {Pärn, E.A. and Edwards, D.J. and Sing, M.C.P.},
  journal = {Automation in Construction},
  volume  = {75},
  pages   = {45-55},
  year    = {2017},
  doi     = {10.1016/j.autcon.2016.12.003}
}

@article{du2007sensorfault,
  title   = {Detection and diagnosis for sensor fault in {HVAC} systems},
  author  = {Du, Zhimin and Jin, Xinqiao},
  journal = {Energy Conversion and Management},
  volume  = {48},
  number  = {3},
  pages   = {693-702},
  year    = {2007},
  doi     = {10.1016/j.enconman.2006.09.023}
}

@article{katipamula2005fddreview,
  title   = {Review Article : Methods for Fault Detection, Diagnostics, and Prognostics for Building Systems—A Review, {Part I}},
  author  = {Katipamula, Srinivas and Brambley, Michael R.},
  journal = {HVAC\&R Research},
  volume  = {11},
  number  = {1},
  pages   = {3-25},
  year    = {2005},
  doi     = {10.1080/10789669.2005.10391123}
}

@article{su2025dtreview,
  title   = {Digital twin and its applications in the construction industry: A state-of-art systematic review},
  author  = {Su, Shuaiming and Zhong, Ray Y. and Jiang, Yishuo},
  journal = {Digital Twin},
  volume  = {2},
  number  = {1},
  year    = {2025},
  doi     = {10.12688/digitaltwin.17664.3}
}

@article{mirnaghi2020fddreview,
  title   = {Fault detection and diagnosis of large-scale {HVAC} systems in buildings using data-driven methods: A comprehensive review},
  author  = {Mirnaghi, Maryam Sadat and Haghighat, Fariborz},
  journal = {Energy and Buildings},
  volume  = {229},
  pages   = {110492},
  year    = {2020},
  doi     = {10.1016/j.enbuild.2020.110492}
}

@article{pei2019dcv,
  title   = {Effect of sensor position on the performance of {CO$_2$}-based demand controlled ventilation},
  author  = {Pei, Gen and Rim, Donghyun and Schiavon, Stefano and Vannucci, Matthew},
  journal = {Energy and Buildings},
  volume  = {202},
  pages   = {109358},
  year    = {2019},
  doi     = {10.1016/j.enbuild.2019.109358}
}

@article{lin2023sitechatbot,
  title   = {Prototyping a Chatbot for Site Managers Using Building Information Modeling ({BIM}) and Natural Language Understanding ({NLU}) Techniques},
  author  = {Lin, Will Y.},
  journal = {Sensors},
  volume  = {23},
  number  = {6},
  pages   = {2942},
  year    = {2023},
  doi     = {10.3390/s23062942}
}

@misc{bommasani2021foundation,
  title  = {On the Opportunities and Risks of Foundation Models},
  author = {Bommasani, Rishi and Hudson, Drew A. and Adeli, Ehsan and others},
  year   = {2021},
  howpublished = {arXiv:2108.07258}
}

@inproceedings{react2023,
  title     = {{ReAct}: Synergizing Reasoning and Acting in Language Models},
  author    = {Yao, Shunyu and Zhao, Jeffrey and Yu, Dian and Du, Nan and Shafran, Izhak and Narasimhan, Karthik and Cao, Yuan},
  booktitle = {Proc.\ 11th Int.\ Conf.\ on Learning Representations (ICLR)},
  year      = {2023}
}

@misc{bats2026,
  title        = {Building {AI} Testbed Suite ({BATS}): {Brick} knowledge graphs, physics-grounded time series and answer keys for nine {DOE} prototype offices},
  author       = {Jung, Wooyoung},
  year         = {2026},
  howpublished = {Zenodo, version 1.0},
  doi          = {10.5281/zenodo.20777376},
  note         = {{CC BY-NC} 4.0}
}

@misc{mortar_hf,
  title        = {Mortar building graphs and time series},
  author       = {Fierro, Gabe},
  howpublished = {Hugging Face datasets \texttt{gtfierro/mortargraphs} and \texttt{gtfierro/mortar}},
  note         = {accessed September 2026}
}

@inproceedings{bts2024,
  title     = {{BTS}: Building timeseries dataset: Empowering large-scale building analytics},
  author    = {Prabowo, Arian and Lin, Xiachong and Razzak, Imran and Xue, Hao and Yap, Emily W. and Amos, Matthew and Salim, Flora D.},
  booktitle = {Advances in Neural Information Processing Systems 37 (NeurIPS 2024)},
  pages     = {133180--133206},
  year      = {2024},
  doi       = {10.52202/079017-4232},
  note      = {Data at \url{https://doi.org/10.6084/m9.figshare.28705559}}
}

@misc{qwen38,
  title        = {Qwen3.8-27B model card},
  author       = {{Qwen Team}},
  howpublished = {Hugging Face / Ollama model repository, tag \texttt{qwen3.8:27b}},
  note         = {accessed September 2026}
}

@misc{gemma4,
  title        = {Gemma 4 Technical Report},
  author       = {{Gemma Team}},
  year         = {2026},
  howpublished = {arXiv:2607.02770}
}

@inproceedings{chen2019gnurl,
  title     = {{Gnu-RL}: A precocial reinforcement learning solution for building {HVAC} control using a differentiable {MPC} policy},
  author    = {Chen, Bingqing and Cai, Zicheng and Berg{\'e}s, Mario},
  booktitle = {Proc.\ 6th ACM Int.\ Conf.\ on Systems for Energy-Efficient Buildings, Cities, and Transportation (BuildSys)},
  pages     = {316--325},
  year      = {2019},
  doi       = {10.1145/3360322.3360849}
}

@article{blum2022mpcfield,
  title   = {Field demonstration and implementation analysis of model predictive control in an office {HVAC} system},
  author  = {Blum, David and Wang, Zhe and Weyandt, Chris and Kim, Donghun and Wetter, Michael and Hong, Tianzhen and Piette, Mary Ann},
  journal = {Applied Energy},
  volume  = {318},
  pages   = {119104},
  year    = {2022},
  doi     = {10.1016/j.apenergy.2022.119104}
}

@inproceedings{fierro2022buildingmotif,
  title     = {Application-driven creation of building metadata models with semantic sufficiency},
  author    = {Fierro, Gabe and Saha, Avijit and Shapinsky, Tobias and Steen, Matthew and Eslinger, Hannah},
  booktitle = {Proc.\ 9th ACM Int.\ Conf.\ on Systems for Energy-Efficient Buildings, Cities, and Transportation (BuildSys)},
  year      = {2022},
  doi       = {10.1145/3563357.3564083}
}

@inproceedings{fierro2020shepherding,
  title     = {Shepherding metadata through the building lifecycle},
  author    = {Fierro, Gabe and Prakash, Anand Krishnan and Mosiman, Cory and Pritoni, Marco and Raftery, Paul and Wetter, Michael and Culler, David E.},
  booktitle = {Proc.\ 7th ACM Int.\ Conf.\ on Systems for Energy-Efficient Buildings, Cities, and Transportation (BuildSys)},
  pages     = {70--79},
  year      = {2020},
  doi       = {10.1145/3408308.3427627}
}

@misc{bimforum2024lod,
  title        = {Level of Development ({LOD}) Specification 2024, {Part I}},
  author       = {{BIMForum}},
  howpublished = {\url{https://bimforum.org/lod}},
  note         = {accessed September 2026}
}

@misc{clockworks2026,
  title        = {Clockworks Analytics: building fault detection and diagnostics software},
  author       = {{Clockworks Analytics}},
  howpublished = {\url{https://clockworksanalytics.com/software/}},
  note         = {product documentation; accessed September 2026}
}

@misc{skyspark2026,
  title        = {{SkySpark} analytics platform},
  author       = {{SkyFoundry}},
  howpublished = {\url{https://skyfoundry.com/product}},
  note         = {product documentation; accessed September 2026}
}

@misc{bara2026,
  title        = {building-ai-readiness: code, results and supplement for a measurement-based {AI}-readiness assessment of buildings},
  author       = {Jung, Wooyoung},
  year         = {2026},
  howpublished = {Zenodo, version 1.0.0},
  doi          = {10.5281/zenodo.23197610},
  note         = {code Apache-2.0; results and {BATS} supplement {CC BY-NC} 4.0}
}

@article{vigna2020sri,
  title   = {Analysis of the Building Smart Readiness Indicator Calculation: A Comparative Case-Study with Two Panels of Experts},
  author  = {Vigna, Ilaria and Pernetti, Roberta and Pernigotto, Giovanni and Gasparella, Andrea},
  journal = {Energies},
  volume  = {13},
  number  = {11},
  pages   = {2796},
  year    = {2020},
  doi     = {10.3390/en13112796}
}

@article{trave2006diagnosability,
  title   = {Diagnosability Analysis Based on Component-Supported Analytical Redundancy Relations},
  author  = {Trav{\'e}-Massuy{\`e}s, Louise and Escobet, Teresa and Olive, Xavier},
  journal = {IEEE Transactions on Systems, Man, and Cybernetics - Part A: Systems and Humans},
  volume  = {36},
  number  = {6},
  pages   = {1146--1160},
  year    = {2006},
  doi     = {10.1109/TSMCA.2006.878984}
}

@article{krysander2008sensorplacement,
  title   = {Sensor Placement for Fault Diagnosis},
  author  = {Krysander, Mattias and Frisk, Erik},
  journal = {IEEE Transactions on Systems, Man, and Cybernetics - Part A: Systems and Humans},
  volume  = {38},
  number  = {6},
  pages   = {1398--1410},
  year    = {2008},
  doi     = {10.1109/TSMCA.2008.2003968}
}

@article{blum2021boptest,
  title   = {Building optimization testing framework ({BOPTEST}) for simulation-based benchmarking of control strategies in buildings},
  author  = {Blum, David and Arroyo, Javier and Huang, Sen and Drgo{\v{n}}a, J{\'a}n and Jorissen, Filip and Walnum, Harald Taxt and Chen, Yan and Benne, Kyle and Vrabie, Draguna and Wetter, Michael and Helsen, Lieve},
  journal = {Journal of Building Performance Simulation},
  volume  = {14},
  number  = {5},
  pages   = {586--610},
  year    = {2021},
  doi     = {10.1080/19401493.2021.1986574}
}

\end{document}